\pdfoutput=1
\documentclass{article}
\usepackage{iclr2026_conference,times}
\usepackage{graphicx}
\usepackage{booktabs}
\usepackage{amsmath,amssymb}
\usepackage{mathtools}
\usepackage{multirow}
\usepackage{adjustbox}
\usepackage[table]{xcolor}   %
\usepackage{tcolorbox}
\tcbuselibrary{skins}
\newtcolorbox{promptbox}[1][]{enhanced,colframe=black!30,colback=white,
  coltitle=black,colbacktitle=black!8,width=\linewidth,arc=2mm,
  auto outer arc,boxrule=0.4pt,left=10pt,right=10pt,
  top=10pt,bottom=10pt,title={#1},
  fonttitle=\bfseries,attach boxed title to top center={yshift=-2mm},
  boxed title style={sharp corners,size=small}}
\usepackage{fontawesome5}    %
\usepackage[section]{placeins}   %
\usepackage{float}
\usepackage{afterpage}

\usepackage{hyperref}
\usepackage{url}
\usepackage{xurl}
\usepackage{letterspace}

\iclrfinalcopy

\title{\protect\lodestartitleicon\protect\kern0.8\fontdimen2\font{\protect\normalfont\protect\textls[220]{LODESTAR}}:\\[0.2ex]
Robust Entropy-Based Answer Selection\\
in Retrieval-Augmented Generation\\
for Question Answering\\[5pt]
\large Directing Frozen-LLM Entropy with a Reinforcement-Learned\\
\large Prompt Polarizer under Misleading Passages}

\author{Hung-Chun Hsu$^{1}$\thanks{Project lead; proposed the methodology, conducted the
experiments, and wrote the manuscript.
\texttt{R10946017@citi.sinica.edu.tw}\enskip $^{\ddagger}$Corresponding author:
\texttt{cjwang@citi.sinica.edu.tw}} \quad
Po-Jen Ko$^{1}$\thanks{Equal contribution.} \quad
Che-Cheng Wu$^{1\dagger}$ \\[3pt]
\textbf{Li-Yang Chang$^{1}$} \quad
\textbf{Chuan-Ju Wang$^{1\ddagger}$} \\[7pt]
$^{1}$Research Center for Information Technology Innovation, Academia Sinica, Taiwan}

\newcommand{\ours}{\textsc{Lodestar}}
\newcommand{\ourstab}{\textsc{Lodestar}~(ours)}
\newcommand{\lodestaricon}{\raisebox{-0.25ex}{\includegraphics[height=1.7ex]{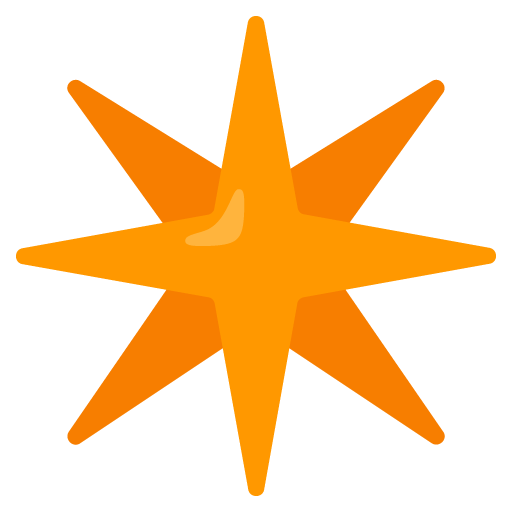}}}
\newcommand{\lodestarinlineicon}{%
  \raisebox{\dimexpr-0.25ex+0.05em\relax}{\includegraphics[height=1.7ex]{figures/assets/noto_2734.png}}\kern0.10em}
\newcommand{\lodestartitleicon}{\raisebox{-0.02ex}{\includegraphics[height=1.65ex]{figures/assets/noto_2734.png}}}
\newcommand{\polar}{\psi^{\star}}   %
\newcommand{\cna}{---}                    %
\definecolor{frozenice}{RGB}{31,119,190}  %
\definecolor{tunedflame}{RGB}{214,96,20}  %
\newcommand{\cfr}{{\color{frozenice}\faSnowflake}}   %
\newcommand{\frozenmark}{{\color{frozenice}\faSnowflake}}
\newcommand{\fresp}{\frozenmark\,respondent}
\newcommand{\cft}{{\color{tunedflame}\faFire}}       %

\begin{document}

\maketitle
\lhead{Preprint}

\begin{abstract}
Predictive-distribution entropy makes a strong selection rule in existing
retrieval-augmented question answering work, which we first verify on our own benchmarks.
Across five QA benchmarks, keeping the candidate answer that a \emph{frozen} respondent
LLM produces with the lowest answer-token entropy lifts mean answer $F_1$ from $0.4769$
to $0.5148$ over the retriever's top-ranked passage without any gold answer.
Yet in this paper we show that this lowest-entropy selection
rule, which prior entropy-based selectors adopt, fails in a specific and consequential way. A misleading passage makes the respondent \emph{confidently
wrong}, driving its entropy down precisely where the signal looks most
trustworthy.
Existing work either scopes its claims away from this failure mode or
documents it without repairing it; we instead show that the failure comes from the
passage that the respondent reads, and the context that passage is read in is an input we
can intervene on.
We therefore
introduce \lodestarinlineicon\ours{}, to the best of our knowledge the first method to
score a text intervention by the uncertainty it induces in a third-party frozen
respondent, compared across one question's candidates, which extends what
the respondent reads instead of retraining what it is.
\ours{} uses reinforcement learning to train, once and offline, a
\emph{polarizer}\textemdash{}a short fixed natural-language string inserted into the
respondent's prompt and never into its weights. Its training labels are built offline
from gold answers and two LLM judges; inference reads neither.
We then evaluate \ours{} comprehensively, running every competing
selector under the same \emph{frozen} respondent and the same retrieved
candidate pool. On $5{,}008$ questions
from those five benchmarks, \ours{} attains
the highest mean $F_1$ of any inference-ready selector, the highest exact
match ($0.4136$), and the highest GPT-4o judge score of the frozen-respondent configurations judged
($0.6435$); its three-seed mean wins all $70$
method-by-dataset cells on $F_1$ against fourteen published configurations, and it takes
the highest macro exact match of any of them, $0.4136$ against the best baseline's
$0.4039$, while remaining paired-significant on $F_1$ against every one.
It raises the same five-benchmark
$F_1$ mean from $0.5148$ to $0.5339$, a further $+3.71\%$. The gain holds on both sides
of the domain split: $0.4643$ to $0.4789$ in-domain on NQ-Open, and $0.5274$ to
$0.5476$ averaged over SQuAD, TriviaQA, EntityQuestions and WebQuestions.
We further ablate the polarizer string and find that it is what makes the respondent read a
misleading passage less often, $26.0\%$ of the time against $30.3\%$.
\end{abstract}

\section{Introduction}
\label{sec:intro}

\paragraph{Misleading passages in retrieval-augmented QA.}
Retrieval-augmented generation (RAG) answers a question from a retrieved corpus rather
than from what the model memorized in training. In open-domain question answering the
answer therefore turns on which passage the model was handed. A typical workflow has two steps:
(1) a retriever returns the corpus passages most similar to the question, and (2) a
respondent LLM, typically frozen, reads those candidates alongside the question and
produces the answer. The trouble is that not all retrieved
passages support the correct answer. Some are topically relevant but factually
misleading, and nothing at inference time marks which is which. A misleading passage is
worse than a useless one, because it supplies a false fact and the respondent's answer
then disagrees with the gold answer. In Table~\ref{tab:misleading-rate} we report how
often that happens, on about a thousand questions from each of five
common QA datasets, where, depending on the retrieval system, between $20.3\%$ and
$35.0\%$ of the retrieved candidates mislead the frozen respondent. What is even worse is that using a
stronger retriever or adding a reranker drives that misleading-passage rate up rather
than down. As a result, deciding which candidate's answer to trust without gold answers
is an active research question in the RAG literature.

\begin{table}[!tp]
\centering
\caption{\textbf{Judged-misleading rate of the retrieved candidate set.} Percentage of a
question's top-$10$ passages the frozen judge (Section~\ref{sec:method}) labels
\textsc{misleading}; \textsc{macro} averages the five columns.}
\label{tab:misleading-rate}
\small
\setlength{\tabcolsep}{4pt}
\begin{adjustbox}{max width=\textwidth}
\begin{tabular}{llcccccc}
\toprule
Candidate pool & Reference & NQ & SQuAD & TriviaQA & EntityQ & WebQ & \textsc{macro} \\
 & & {\footnotesize $n{=}1{,}000$} & {\footnotesize $n{=}1{,}000$} & {\footnotesize $n{=}1{,}000$} & {\footnotesize $n{=}1{,}008$} & {\footnotesize $n{=}1{,}000$} & {\footnotesize $n{=}5{,}008$} \\
\midrule
\rowcolor{gray!12}
Contriever & {\footnotesize\citep{izacard2022contriever}} & $24.2$ & $21.8$ & $11.8$ & $22.3$ & $21.4$ & $20.3$ \\
\quad $+$ bge-reranker-v2-gemma & {\footnotesize\citep{chen2024bgem3}} & $32.5$ & $31.5$ & $17.5$ & $34.6$ & $30.7$ & $29.4$ \\
\quad $+$ Qwen3-Reranker-8B & {\footnotesize\citep{zhang2025qwen3emb}} & $38.6$ & $33.8$ & $18.7$ & $39.3$ & $32.6$ & $32.6$ \\
\midrule
\rowcolor{gray!12}
bge-m3 & {\footnotesize\citep{chen2024bgem3}} & $33.4$ & $27.2$ & $16.0$ & $36.3$ & $31.4$ & $28.9$ \\
\quad $+$ bge-reranker-v2-gemma & {\footnotesize\citep{chen2024bgem3}} & $33.6$ & $32.8$ & $18.6$ & $41.2$ & $34.1$ & $32.1$ \\
\quad $+$ Qwen3-Reranker-8B & {\footnotesize\citep{zhang2025qwen3emb}} & $39.5$ & $34.8$ & $19.8$ & $45.6$ & $35.5$ & $35.0$ \\
\midrule
Qwen3-Embedding-8B & {\footnotesize\citep{zhang2025qwen3emb}} & $35.1$ & $30.9$ & $17.7$ & $34.5$ & $30.2$ & $29.7$ \\
\bottomrule
\end{tabular}
\end{adjustbox}
\par\vspace{-8pt}
\end{table}

\begin{figure}[!hb]
\centering
\includegraphics[width=\textwidth]{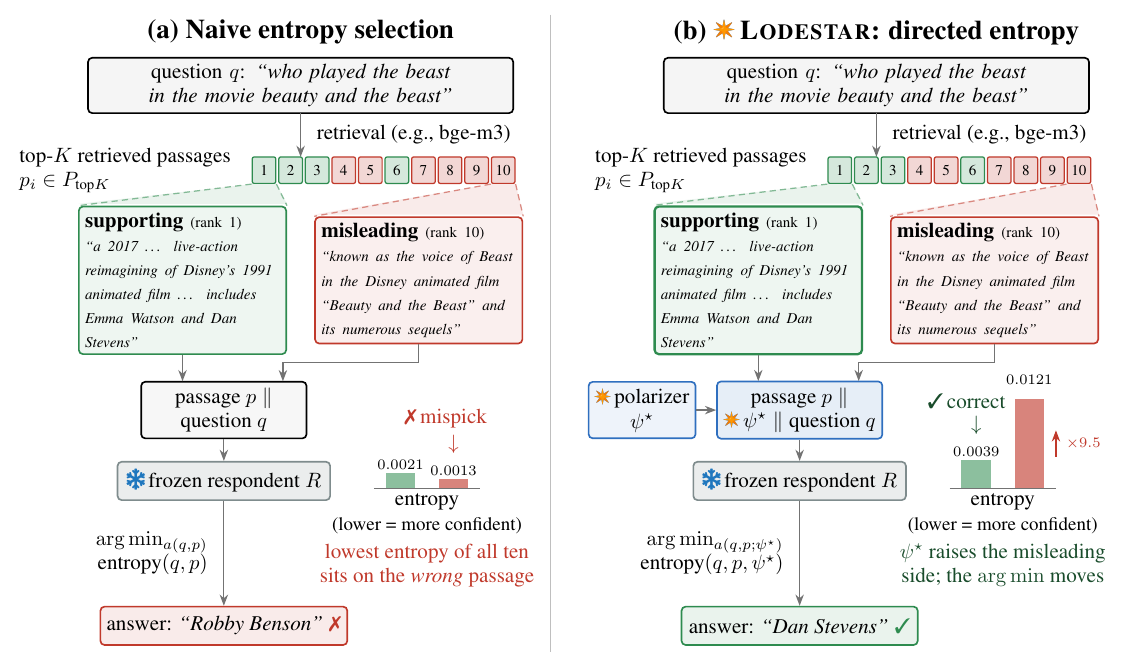}
\caption{\textbf{Why minimum-entropy selection is misled, and what \ours{} changes.}
(a)~The misleading passage makes the frozen respondent become
\emph{confidently} wrong: the respondent's entropy is lowest on that passage
($0.0013$, the minimum over all ten, against the supporting passage's $0.0021$), and
the lowest-entropy rule keeps the answer that passage induces, which is wrong.
(b)~\ours{} inserts one learned polarizer $\polar$ between each passage and the
question, optimized to raise the respondent's entropy on misleading passages while
leaving supporting ones near the entropy floor. The misleading passage rises
$9.5\times$ on the unrounded probes, to $0.0121$, while the supporting one rises only to
$0.0039$, so the same
lowest-entropy rule now selects the supporting passage and the answer is correct. No
gold answers are used at selection time.}
\label{fig:overview}
\end{figure}

\paragraph{Entropy selection and where it fails.}
Entropy-based selection is an effective way to choose among the candidates
\underline{without} gold answers, taking the entropy of the answer tokens produced under each candidate as that
candidate's score. The rule is then to keep the answer the respondent produces with the
least uncertainty (i.e., the lowest entropy). Across the same five QA datasets, even
the simplest entropy signal lifts mean answer $F_1$ from $0.4769$ to $0.5148$ over simply
taking the retriever's top-ranked passage. Prior work computes this confidence in several
forms, all from the respondent's own distribution: entropy over sampled generations
clustered by meaning \citep{farquhar2024semantic}, over the generation trajectory
\citep{epr2026,clehe2025}, and over a short probe rollout \citep{song2026igp}.
Yet the \emph{confidently-wrong problem} is where
that effectiveness ends. The passage this rule selects still yields an answer with zero exact match
$59.6\%$ of the time. That low uncertainty does not imply correctness is documented
\citep{anatomy2026,semanticenergy2025}, and so is the finding that stronger retrieval
systematically lowers predictive entropy \citep{soudani2025axioms}. Neither result says
which of one question's own candidates to trust; that is where the direction reverses, and
within a single question a misleading passage lowers the respondent's entropy
(Figure~\ref{fig:overview}a).
The failure is more often sidestepped than met (Section~\ref{sec:related}), leaving the
\emph{confidently-wrong problem} unsolved.

\paragraph{From measured to directed entropy.}
To proactively address the \emph{confidently-wrong problem} we introduce \ours{}
(Learned Orientation of Directed Entropy,
Steering Trustworthy Answer Retrieval), which leaves the respondent frozen and learns
instead one short natural-language \emph{polarizer} $\polar$, inserted after the passage $p$
and before the question $q$. The prompt then reads
$[\,p\,;\,\polar\,;\,q\,]$.
Across the same five datasets, \ours{}
further lifts mean answer $F_1$ from entropy selection's $0.5148$ to $0.5339$
(Section~\ref{sec:main}). Trained once and offline by reinforcement learning, $\polar$ is
optimized against a reward that is not answer correctness but the \emph{within-question}
separation of the frozen respondent's entropy, raised on misleading passages and kept low
on supporting ones (Figure~\ref{fig:overview}b).
We call the resulting signal \emph{directed}
entropy, because the polarizer steers the respondent's entropy in a chosen direction
rather than leaving it to be passively measured. \ours{} runs at inference with no extra model, no
sampling and no supervision; the only thing it keeps from training is the polarizer
$\polar$, one fixed string, enough by itself to outperform every published method the
paper tests (Section~\ref{sec:main}). Our
contributions can be summarized as follows:
\begin{enumerate}
\itemsep2pt
\item \textbf{We show that ranking a question's candidates by entropy alone is unreliable,
and repair it with one learned string.} Within a single question, the respondent's entropy
does not reliably separate misleading passages from supporting ones
(Section~\ref{sec:method}); one learned string is enough to make it separate them. Ranking
by entropy alone reads a misleading passage more often on average than drawing one of
the ten at random, $30.3\%$ against the pool's own $28.9\%$. One learned string puts the selection
below that floor on all five benchmarks, $26.0\%$ macro-averaged
(Tables~\ref{tab:misleading-rate} and~\ref{tab:ablation}).
\item \textbf{We benchmark fourteen published methods re-purposed as selectors; \ours{}
leads them on every metric it is measured on.} Fourteen published configurations, each run
on identical pools with an identical frozen respondent under the strongest setting its
released code or paper supports (Section~\ref{sec:settings}; per-row deviations in
Appendix~\ref{app:baselines}), spanning prompt-search
optimizers, uncertainty signals scored from the respondent's own output, and trained
rerankers that never see it. \ours{} attains the highest mean answer $F_1$ ($0.5339$) and
exact match ($0.4136$) of any of them, and the highest GPT-4o judge score of the frozen-respondent
configurations judged ($0.6435$); its $F_1$
lead is paired-significant against every configuration tested
(Appendix~\ref{app:significance}).
\item \textbf{We map the polarizer over all nine train--inference pairs of three frozen
respondents, Llama-3.1-8B, Qwen2.5-7B and Qwen3.5-9B; every diagonal helps, though the
diagonal is not always where it helps most.}
The better entropy signal tracks the model family: first-token $H_1$
on Llama, all-token on both Qwen models.
What holds across all three is the polarizer's effect. %
Section~\ref{sec:ablation} reports the cross-respondent comparison.
\end{enumerate}

\section{Related Work}
\label{sec:related}

\paragraph{The confidently-wrong problem.}
That low uncertainty does not imply correctness is documented in the literature
\citep{anatomy2026,
semanticenergy2025}, and \citet{soudani2025axioms} establish that stronger retrieval
systematically lowers predictive entropy. As we show in Table~\ref{tab:misleading-rate},
retrieving with bge-m3 puts a higher average share of misleading passages in the
top-$10$ than retrieving with Contriever, $28.9\%$ against $20.3\%$. Moreover, on our
five pools the respondent's mean first-token entropy is
$1.22$ nats\footnote{The unit of entropy taken with the natural logarithm.} on
judged-misleading
candidates against $1.33$ on the rest of the candidate set. A selector that keeps the
lowest-entropy candidate is therefore drawn toward misleading passages and the
incorrect answers they induce.
Yet what a selector needs is directional. Reading a misleading passage should
leave the respondent less certain, its answer-token entropy the higher one. Neither
prior result meets this need. Both describe how entropy behaves across questions, while
a selector only compares the candidates of one question, a gap \ours{} closes by
training its polarizer to make misleading passages carry the higher entropy inside
each question (Section~\ref{sec:method}).

\paragraph{Uncertainty estimation for generation.}
Predictive entropy and its refinements are the standard estimators for measuring how
uncertain a language model is about its answer. Semantic entropy
\citep{farquhar2024semantic} clusters
sampled generations by entailment before computing entropy; EigenScore
\citep{chen2024inside} and SeaKR \citep{yao2025seakr} read the dispersion (e.g.,
covariance eigenvalues, Gram determinant) of internal
states across samples; EPR \citep{epr2026} and \textsc{CLeHe} \citep{clehe2025} score
the generation trajectory, the step-by-step entropy of the answer as the respondent
decodes it. Three further signals reach this task from outside
hallucination detection. Min-K\% Prob \citep{shi2024mink} and its
vocabulary-normalized successor Min-K\%++ \citep{zhang2025minkpp} were introduced to
test whether a text was part of a model's training data. A text the model has seen
keeps even its least-likely tokens probable. Re-purposed as selectors, they score the
answer conditioned on each passage and keep the passage whose answer scores highest.
SPS \citep{sps2026}
scores a passage from its representation alone, without seeing the question. None of
these signals changes the input that produced it. Yet with a frozen respondent the
input is the only part of the system a selector is free to change; \ours{} uses
exactly that freedom, learning one string that moves the respondent's entropy instead
of merely reading it.

\paragraph{Entropy-based answer selection.}
Uncertainty also governs the upstream decision of \emph{whether} to retrieve at all
\citep{moskvoretskii2025adarague}; our question is which candidate's answer to keep for
the frozen respondent once retrieval has happened. The closest prior method that does
select is IGP \citep{song2026igp}. It ranks the retrieved candidates $p \in P(q)$ by the
information gain $\mathrm{IG}(p,q) = \bar{H}_{L}(q) - \bar{H}_{L}(q,p)$ each delivers to a
frozen respondent, with $\bar{H}_{L}(q,p)$ the mean normalized entropy of the $L$ tokens
that compose a short probe answer and $\bar{H}_{L}(q)$ the same entropy with no passage
in context. Because $\bar{H}_{L}(q)$ is constant once the question is fixed, it cancels
within a question, and the rule reduces to a minimum-entropy selector over the answers
$a(q,p)$ the candidates induce:
\begin{equation}
\underbrace{\hat{a}^{\mathrm{IGP}}}_{\mathclap{\text{selected answer}}} \;=\;
\arg\max_{a(q,p),\ p \in P(q)} \underbrace{\mathrm{IG}(p,q)}_{\text{information gain}}
\;=\;
\arg\min_{a(q,p),\ p \in P(q)} \underbrace{\bar{H}_{L}(q,p)}_{\mathclap{\text{measured entropy}}}.
\label{eq:igp}
\end{equation}
A second strategy learns the decision rather than
scoring it, and this does not by itself escape the confidently-wrong problem either. MBA-RAG
\citep{tang2025mbarag} learns a bandit policy conditioned on the question alone; a
policy that never reads the candidates can decide how to retrieve for a question, but
not which of that question's candidates to trust (Section~\ref{sec:main}).

\paragraph{Learned guidance for retrieval-augmented answering.}
A line of work trains a model to emit text that guides how a retrieval-augmented system
answers.
Self-RAG
\citep{asai2024selfrag} trains a model to generate reflection tokens;
GainRAG \citep{gainrag2025} aligns retriever and respondent through a perplexity-based
gain signal whose teacher reads the gold answer; CTRL-RAG \citep{ctrlrag2026} uses
reinforcement learning to make generation more context-faithful; and CRITIC-R1
\citep{criticr12026} leaves the generator frozen and trains a critic instead, rewarding it
for aligning its verdicts with a teacher's judgement of the answer.
What they share is where the training target is read: from the text itself, whether that
is an answer's correctness, its groundedness in the passage, the passage's relevance, or
the quality of a critique of the trajectory.
We stress that Self-RAG and CTRL-RAG train the answering
model itself. Their respondent is not frozen, so they solve a different problem from
ours, improving the model that answers rather than the evidence a fixed model answers
from.
\ours{} changes only the training
signal. To the best of our knowledge, no prior work scores a text intervention by the
uncertainty it induces in a third-party frozen respondent, compared across the candidates
of one question.

\section{Methodology}
\label{sec:method}

\paragraph{The problem: answer selection for a frozen respondent.}
A frozen respondent $R$ is only as good as the single passage (possibly misleading, possibly
supporting) it reads. That choice decides whether its answer is correct. Given a question
$q$ and a retrieved candidate set $P(q) = \{p_1,\dots,p_K\}$, each candidate read alone
leads $R$ to one candidate answer, so the pool induces $K$ answers to choose among.

\paragraph{The signal: within-question entropy separation.}
Within a single question, a misleading passage \emph{lowers} the respondent's entropy;
selection by answer-token entropy is therefore drawn to the wrong answers $R$ produces
confidently from misleading passages, over the less confident but correct ones.
We denote $\bar{H}_{L}(q,p)$ as the mean normalized entropy of the $L$ tokens of the answer
$R$ generates from passage $p$ alone, and $a(q,p)$ as that answer. The answer
selector keeps $\arg\min_{a(q,p),\ p \in P(q)} \bar{H}_{L}(q,p)$, which is the rule
Equation~\ref{eq:igp} showed IGP reduces to.
Let $\mathrm{Mis}(q)$ and $\mathrm{Sup}(q)$ be the passages of question $q$ labelled
misleading and supporting. The quantity \ours{} acts on is the within-question separation
between the entropies $R$ produces on $\mathrm{Mis}(q)$ and on $\mathrm{Sup}(q)$, which the
polarizer is trained to drive positive inside each question:
\begin{equation}
\Delta_{\mathrm{Mis}-\mathrm{Sup}} \bar{H}_{L}(q) \;=\; \operatorname*{mean}_{p\in\mathrm{Mis}(q)} \bar{H}_{L}(q,p)
\;-\; \operatorname*{mean}_{p\in\mathrm{Sup}(q)} \bar{H}_{L}(q,p).
\label{eq:sep}
\end{equation}

\paragraph{The intervention: directed entropy.}
\ours{}'s methodological contribution is the \emph{RL-learned polarizer}, one fixed
natural-language string $\polar$ that turns the respondent's entropy from a quantity we
passively measure into one we intervene on.
\ours{} leaves $R$ frozen and inserts $\polar$ after the passage $p$ and before the
question $q$ (the prompt is given in Appendix~\ref{app:grpo}). We call the result,
$\bar{H}_{L}(q,p,\polar)$, the \emph{directed entropy}. The polarizer $\polar$ is
optimized to move it with an intended sign, upward on misleading passages, unchanged on
supporting ones. Under directed entropy the answer selector differs from
Equation~\ref{eq:igp} by one symbol, the entropy it ranks now computed under
$\polar$:
\begin{equation}
\underbrace{\hat{a}}_{\mathclap{\text{selected answer}}} \;=\; \arg\min_{a(q,p;\polar),\ p \in P(q)} \underbrace{\bar{H}_{L}(q, p, \polar)}_{\text{directed entropy}}.
\label{eq:select}
\end{equation}

\paragraph{Training objective.}
The optimal polarizer $\polar$ is the string that maximizes the within-question
separation its insertion induces. A generator policy proposes candidate polarizers from
a single fixed prompt that contains neither labels nor gold
answers,\footnote{The converged strings are reported in
Appendix~\ref{app:polarizertext}, where independent seeds are shown to agree on what to
say even though the reward never scores wording.} and is trained on
their reward by GRPO \citep{shao2024deepseekmath}.
Each candidate polarizer $\psi$ is
credited with the separation of Equation~\ref{eq:sep} evaluated under $\psi$,
baseline-corrected against the same question's natural separation:
\begin{equation}
\begin{gathered}
\mathcal{S}(\psi) \;=\; \widehat{\mathbb{E}}_{q \in B}\Big[\,\underbrace{\Delta_{\mathrm{Mis}-\mathrm{Sup}} \bar{H}_{L}(q;\psi)}_{\text{separation with polarizer } \psi \text{ inserted}} \;-\; \underbrace{\Delta_{\mathrm{Mis}-\mathrm{Sup}} \bar{H}_{L}(q)}_{\text{the natural separation, Equation~\ref{eq:sep}}}\,\Big],\\
\text{with}\quad \Delta_{\mathrm{Mis}-\mathrm{Sup}} \bar{H}_{L}(q;\psi) \;=\; \operatorname*{mean}_{p\in\mathrm{Mis}(q)} \bar{H}_{L}(q,p,\psi) \;-\; \operatorname*{mean}_{p\in\mathrm{Sup}(q)} \bar{H}_{L}(q,p,\psi),
\end{gathered}
\label{eq:reward}
\end{equation}
where $\widehat{\mathbb{E}}_{q\in B}$ averages over the question set $B$ scored at that step,
whose labelled passages are sampled from the top-$K$ retrieved. $\mathcal{S}(\psi)$
is the polarizer's reward, zero whenever $\psi$ changes nothing and larger being better.
The reward functional never reads an answer, gold or generated. The partition $\mathrm{Mis}(q)/\mathrm{Sup}(q)$ that the reward contrasts is
nonetheless built from gold- and judge-derived labels, so
gold-freedom is a property of inference, not of training-data construction.
Appendix~\ref{app:grpo} states the GRPO objective and its advantage in full, together
with every setting.

\paragraph{Passage labels.}
Two LLM judges from different model families
\citep{zheng2023judging,verga2024juries,thomas2024searcher}, gpt-oss-120b \citep{openai2025gptoss} and
Qwen2.5-72B-Instruct \citep{qwen2024qwen25}, read each passage independently. A passage
counts as \textsc{misleading} only when both agree that it is, as \textsc{supporting} when
the respondent's answer from it exactly matches a gold answer after normalization, and as
\textsc{neutral} otherwise; the rare passage meeting both criteria is routed to
\textsc{neutral}. On
the $5{,}000$ NQ-Open questions from which the training pool is drawn they agree with
Cohen's $\kappa = 0.675$, and on a held-out audit set an independent third-party judge
returns their consensus label $96\%$ of the time (Appendix~\ref{app:judge}).

\paragraph{Inference.}
Inference adds no model, no sampling, no supervision and no passage label. The respondent
is run once per candidate to obtain $\bar{H}_{L}(q,p,\polar)$, and Equation~\ref{eq:select}
returns the argmin. $\polar$ stays in the prompt while that answer is decoded, which is
why Equation~\ref{eq:select} selects among $a(q,p;\polar)$ rather than $a(q,p)$. The only
artifact training carries forward is $\polar$, one
category-level string, and selection reads nothing but the respondent's own entropy.

\section{Experimental Settings}
\label{sec:settings}

\subsection{Benchmark}
\label{sec:settings-benchmark}

\paragraph{Datasets and evaluation pools.}
We evaluate \ours{} and all baselines on five open-domain QA datasets: \emph{in-domain},
Natural Questions \citep{kwiatkowski2019nq}; \emph{out-of-domain}, SQuAD, TriviaQA,
EntityQuestions and WebQuestions \citep{rajpurkar2016squad, joshi2017triviaqa,
sciavolino2021entityquestions, berant2013webquestions}, none of whose training data is
ever seen.
Evaluation pools are fixed in advance and shared by every method: $n\!=\!1{,}000$ each
except EntityQuestions at $n\!=\!1{,}008$, each with the top ten bge-m3
\citep{chen2024bgem3} candidates from one shared Wikipedia index.

\paragraph{Frozen respondent and answer evaluation protocol.}
The frozen respondent is Llama-3.1-8B-Instruct \citep{grattafiori2024llama3}, its
parameters never updated.\footnote{We additionally run the cross-respondent comparison across two
model generations, Qwen2.5-7B and Qwen3.5-9B, training a polarizer for each
respondent and evaluating every pairing both in-domain and out-of-domain; every respondent
improves under its own polarizer on both sides of the split (Section~\ref{sec:ablation}).}
It answers greedily from the selected passage alone, under one shared answering template
that carries $\polar$ for \ours{} and nothing extra for every other configuration.
Every $F_1$ in this paper scores that generated answer against the gold answer(s); exact
match (EM) does the same after normalization. An LLM judge, GPT-4o \citep{openai2024gpt4o},
additionally rates whether the generated answer is equivalent to any gold answer, under a
judging prompt adopted verbatim from \citet{farquhar2024semantic}. Appendix~\ref{app:em}
carries the exact-match table and full judge protocol, and Appendix~\ref{app:judge} reports
the judge's numbers per dataset.

\subsection{Polarizer training}
\label{sec:settings-training}

\paragraph{Policy, data and budget.}
We adopt Qwen3-4B-Instruct \citep{qwen2025qwen3} as the polarizer policy, trading
generation quality against training cost at a scale we can afford to run three times, and
train it with GRPO (Appendix~\ref{app:grpo}). The polarizer's training questions are the full
$1{,}939$-question NQ-Open training pool, disjoint by construction from
every evaluation pool. The reported $\polar$, and every
polarizer reported in this paper, is the \emph{final} checkpoint of a completed,
fixed $300$-step budget.

\paragraph{Entropy estimator.}
\label{sec:settings-estimator}
Reproducing IGP, whose gain term averages entropy over all $L$ answer tokens,
we found that scoring the first answer token alone is no worse on the macro mean under a
shared answering prompt ($0.5148$ against $0.5001$ mean answer $F_1$; Appendix~\ref{par:igp-aligned}).
$L{=}1$ is therefore the setting \ours{} uses, in its training reward and in its selector
alike. The choice has precedent on both counts: reading a selection score from the first
decoded position alone is established in neural reranking, where
\citet{gangireddy2024first} take a full listwise ordering from the logits of the first
generated identifier; and approximating a multi-token entropy by its first token is a
standing convention, one \citet{clark2025firsttoken} test directly and caution against.
The measurement above asks their question of this task, and answers it.
Each baseline keeps the estimator its own paper specifies.
$\bar{H}_{L}$ becomes the
respondent's first-token entropy $H_1$, and the decoding a candidate needs to be scored
collapses from $L$ tokens to one, an $L$-fold saving on every candidate of every question.

\subsection{Baselines}
\label{sec:settings-baselines}

\ifdefined\emjudgecolwd\else\newlength{\emjudgecolwd}\fi
\ifdefined\emheadwd\else\newlength{\emheadwd}\fi
\settowidth{\emjudgecolwd}{\small$\mathit{LLM\text{-}Judge}$ $\uparrow$}
\settowidth{\emheadwd}{\small\textbf{Answer $EM$} $\uparrow$}
\ifdim\emheadwd>\emjudgecolwd\setlength{\emjudgecolwd}{\emheadwd}\fi
\ifdefined\fonecolwd\else\newlength{\fonecolwd}\fi
\settowidth{\fonecolwd}{\small In-domain}
\makeatletter
\ifdefined\jrhairline\else
  \newsavebox{\jrbox}
  \newif\ifjrready
  \def\jrhairlinewd{0.4pt}
  \def\jrsavepos#1{\pdfsavepos
    \write\@auxout{\string\gdef\string#1{\the\pdflastypos}}}
  \def\jrhairlinevoid{\vrule width\jrhairlinewd height\z@ depth\z@\relax}
  \def\jrhairlinepiece{\textcolor{black!20}{\vrule width\jrhairlinewd}}
  \def\jrseg#1#2{\rlap{\raise\dimexpr#1sp-\jrposR sp\relax
    \hbox{\vrule width\jrhairlinewd height\dimexpr#2sp-#1sp\relax depth\z@}}}
  \def\jrsegments{%
    \jrseg\jrposHb\jrposHa
    \jrseg\jrposBa\jrposTa \jrseg\jrposBb\jrposTb \jrseg\jrposBc\jrposTc
    \jrseg\jrposBd\jrposTd \jrseg\jrposBe\jrposTe}
  \def\jrhairlinedraw{\jrsavepos\jrposR
    \ifjrready
      \sbox\jrbox{\textcolor{black!20}{\jrsegments\kern\jrhairlinewd}}%
      \ht\jrbox\z@ \dp\jrbox\z@ \usebox\jrbox
    \else\jrhairlinepiece\fi}
  \def\jrcheckspan#1#2{%
    \ifdim\dimexpr#2sp-#1sp\relax<0.5pt \jrreadyfalse\fi
    \ifdim\dimexpr#2sp-#1sp\relax>400pt \jrreadyfalse\fi}
  \def\jrsetup{%
    \jrreadytrue
    \@for\jrtmp:=Ha,Hb,Ta,Ba,Tb,Bb,Tc,Bc,Td,Bd,Te,Be,R\do
      {\@ifundefined{jrpos\jrtmp}{\jrreadyfalse}{}}%
    \ifjrready
      \jrcheckspan\jrposHb\jrposHa
      \jrcheckspan\jrposBa\jrposTa \jrcheckspan\jrposBb\jrposTb
      \jrcheckspan\jrposBc\jrposTc \jrcheckspan\jrposBd\jrposTd
      \jrcheckspan\jrposBe\jrposTe
    \fi
    \ifjrready
      \global\let\jrhairline\jrhairlinevoid
    \else
      \global\let\jrhairline\jrhairlinepiece
    \fi}
\fi
\makeatother
\jrsetup
\makeatletter
\ifdefined\sfp@cell\else
  \newlength{\sfp@cell}    %
  \newlength{\sfp@colwd}   %
  \newlength{\sfp@rowht}   %
  \newsavebox{\sfp@art}\newsavebox{\sfp@probe}
  \newcommand{\sfp@ug}[1]{\makebox[\sfp@cell][c]{#1}}
  \newcommand{\sfp@ding}[1]{{\usefont{U}{ding}{m}{n}\symbol{#1}}}
  \definecolor{sfpink}{gray}{0.72}
  \def\sfp@map#1{%
    \if#1f\sfp@ding{'065}\else
    \if#1o\sfp@ding{'066}\else
    \if#1F\sfp@ding{'070}\else
    \if#1s\ensuremath{\star}\else
    \if#1.\textperiodcentered\else\fi\fi\fi\fi\fi}
  \def\sfp@stop{\sfp@stop}
  \def\sfp@scan#1{\ifx#1\sfp@stop\let\sfp@next\relax\else
    \let\sfp@next\sfp@scan \sfp@ug{\sfp@map{#1}}\fi\sfp@next}
  \newcommand{\sfp@row}[1]{\hbox{\sfp@scan#1\sfp@stop}}
  \newcommand{\sfp@build}{%
    \setbox\sfp@art=\vbox{\begingroup
      \scriptsize\baselineskip=14.89pt\relax\color{sfpink}%
      \sfp@row{--f------.------o-------.--------f}%
      \sfp@row{-----.------s--------F-------.------.}%
      \sfp@row{-s------.---------f-------.-------o}%
      \sfp@row{----.---------.-------s-------.-----f}%
      \endgroup}%
    \setbox\sfp@art=\hbox{$\vcenter{\box\sfp@art}$}}
  \newcommand{\sfpfit}{%
    \sbox{\sfp@probe}{\resizebox{\sfp@colwd}{!}{\usebox{\sfp@art}}}%
    \ifdim\dimexpr\ht\sfp@probe+\dp\sfp@probe\relax>\sfp@rowht
      \resizebox*{!}{\sfp@rowht}{\usebox{\sfp@art}}%
    \else
      \usebox{\sfp@probe}%
    \fi}
\fi
\settowidth{\sfp@cell}{\ttfamily\scriptsize M}
\settowidth{\sfp@colwd}{\small{\footnotesize\emph{mean over ten bge-m3 candidates}}}
\begingroup\small\global\sfp@rowht=\dimexpr4\baselineskip\relax\endgroup
\sfp@build
\makeatother
\begin{table}[!t]
\centering
\caption{\textbf{Passage-selection $F_1$ with a frozen Llama-3.1-8B-Instruct
respondent.} All configurations select one of the same ten bge-m3 candidates and answer with the
same frozen respondent, so the rows differ only in how that passage is chosen.
\emph{Mean}, $EM$ and the LLM-judge column are five-pool macro means of $F_1$, of exact
match (Appendix~\ref{app:em}) and of a GPT-4o answer-equivalence verdict
(Appendix~\ref{app:judge}). The \ours{} row is the mean of the three runs listed beneath
it in all three.}
\label{tab:main}
\setlength{\tabcolsep}{3.5pt}
\small
\begin{adjustbox}{max width=\textwidth}
\begin{tabular}{llw{c}{\fonecolwd}w{c}{\fonecolwd}w{c}{\fonecolwd}w{c}{\fonecolwd}w{c}{\fonecolwd}w{c}{\fonecolwd}@{\hspace{1.6\tabcolsep}\jrhairline\hspace{1.6\tabcolsep}}w{c}{\emjudgecolwd}@{\hspace{1.6\tabcolsep}\jrhairline\hspace{1.6\tabcolsep}}w{c}{\emjudgecolwd}}
\toprule
\noalign{\jrsavepos\jrposHa}%
 &
   & \multicolumn{6}{c}{\textbf{Answer $F_1$} $\uparrow$}
   & \textbf{Answer $EM$} $\uparrow$ & $\mathit{LLM\text{-}Judge}$ $\uparrow$ \\  %
\cmidrule(lr{\dimexpr\cmidrulekern+2.2\tabcolsep+0.4pt\relax}){3-8}%
\cmidrule(lr{\dimexpr\cmidrulekern+3.2\tabcolsep+0.4pt\relax}){9-9}%
\cmidrule(lr{\dimexpr\cmidrulekern+\tabcolsep\relax}){10-10}
\multirow{2}{*}{Method} & \multirow{2}{*}{Reference} &
  In-domain & \multicolumn{4}{c}{Out-of-domain} & \multirow{2}{*}{Mean}
  & \raisebox{-15.87231pt}[0pt][0pt]{\footnotesize\shortstack{Mean\\[-0.81165pt]full table in\\[-0.81165pt]Appendix~\ref{app:em}}}
  & \raisebox{-15.87231pt}[0pt][0pt]{\footnotesize\shortstack{Mean\\[-0.81165pt]full table in\\[-0.81165pt]Appendix~\ref{app:judge}}} \\
\cmidrule(lr){3-3} \cmidrule(lr){4-7}
  & & NQ & SQuAD & TriviaQA & EntityQ & WebQ &
  & & \\
\noalign{\jrsavepos\jrposHb}%
\midrule
\rowcolor{gray!12}
\multicolumn{10}{l}{\textsc{RL-learned polarizer}} \\
\noalign{\jrsavepos\jrposTa}%
\addlinespace[1pt]
\textbf{\lodestaricon~\ourstab{}} & \multirow{4}{*}{\sfpfit} & \textbf{0.4789} & \textbf{0.5968} & \textbf{0.7181} & \textbf{0.4839} & \textbf{0.3917} & \textbf{0.5339} & \textbf{0.4136} & \textbf{0.6435} \\
\hspace{1em} -- training-data seed 42 & & 0.4783 & 0.5916 & 0.7163 & 0.4943 & 0.3908 & 0.5343 & 0.4148 & 0.6420 \\
\hspace{1em} -- training-data seed 43 & & 0.4707 & 0.5954 & 0.7215 & 0.4661 & 0.3877 & 0.5283 & 0.4107 & 0.6365 \\
\hspace{1em} -- training-data seed 44 & & 0.4877 & 0.6033 & 0.7164 & 0.4912 & 0.3966 & 0.5390 & 0.4152 & 0.6520 \\
\noalign{\jrsavepos\jrposBa}%
\midrule
\rowcolor{gray!12}
\multicolumn{10}{l}{\textsc{Published signals re-purposed as selectors}} \\
\noalign{\jrsavepos\jrposTb}%
\addlinespace[1pt]
\hspace{1em}Semantic entropy & {\footnotesize\citep{farquhar2024semantic}} & 0.4616 & 0.5650 & 0.6997 & 0.4430 & 0.3798 & 0.5098 & 0.4039 & 0.6001 \\
\hspace{1em}Self-RAG (reflection) & {\footnotesize\citep{asai2024selfrag}} & 0.4474 & 0.5585 & 0.6856 & 0.4640 & 0.3805 & 0.5072 & 0.3833 & 0.6212 \\
\hspace{1em}SeaKR & {\footnotesize\citep{yao2025seakr}} & 0.4462 & 0.5708 & 0.6944 & 0.4386 & 0.3748 & 0.5050 & 0.3959 & 0.6006 \\
\hspace{1em}CLeHe & {\footnotesize\citep{clehe2025}} & 0.4346 & 0.5749 & 0.6817 & 0.4377 & 0.3719 & 0.5002 & 0.3783 & 0.6063 \\
\hspace{1em}EPR & {\footnotesize\citep{epr2026}} & 0.4333 & 0.5643 & 0.6763 & 0.4367 & 0.3732 & 0.4968 & 0.3753 & 0.6025 \\
\hspace{1em}Min-K\% Prob & {\footnotesize\citep{shi2024mink}} & 0.4345 & 0.5466 & 0.6909 & 0.4210 & 0.3693 & 0.4924 & 0.3797 & 0.5886 \\
\hspace{1em}IGP & {\footnotesize\citep{song2026igp}} & 0.4307 & 0.5169 & 0.6847 & 0.4367 & 0.3531 & 0.4844 & 0.3607 & 0.5935 \\
\hspace{1em}EigenScore & {\footnotesize\citep{chen2024inside}} & 0.4163 & 0.5437 & 0.6711 & 0.4160 & 0.3624 & 0.4819 & 0.3711 & 0.5754 \\
\hspace{1em}SPS & {\footnotesize\citep{sps2026}} & 0.3005 & 0.2606 & 0.5688 & 0.2457 & 0.3146 & 0.3380 & 0.2439 & 0.4075 \\
\hspace{1em}Min-K\%++$^{\P}$ & {\footnotesize\citep{zhang2025minkpp}} & 0.2186 & 0.1947 & 0.5030 & 0.1872 & 0.2586 & 0.2724 & 0.1994 & 0.3250 \\
\noalign{\jrsavepos\jrposBb}%
\midrule
\rowcolor{gray!12}
\multicolumn{10}{l}{\textsc{Trained reranker (respondent-free)}} \\
\noalign{\jrsavepos\jrposTc}%
\addlinespace[1pt]
\hspace{1em}GainRAG & {\footnotesize\citep{gainrag2025}} & 0.4433 & 0.5057 & 0.6836 & 0.4675 & 0.3753 & 0.4951 & 0.3707 & 0.6054 \\
\hspace{1em}MBA-RAG & {\footnotesize\citep{tang2025mbarag}} & 0.4191 & 0.5245 & 0.6513 & 0.4384 & 0.3491 & 0.4765 & 0.3591 & 0.5747 \\
\hspace{1em}CRITIC-R1 & {\footnotesize\citep{criticr12026}} & 0.4085 & 0.4987 & 0.6402 & 0.3980 & 0.3674 & 0.4626 & 0.3378 & 0.5712 \\
\noalign{\jrsavepos\jrposBc}%
\midrule
\rowcolor{gray!12}
\multicolumn{10}{l}{\textsc{Search-based guidance}} \\
\noalign{\jrsavepos\jrposTd}%
\addlinespace[1pt]
\hspace{1em}GEPA & {\footnotesize\citep{agrawal2025gepa}} & 0.4631 & 0.5876 & 0.7081 & 0.4446 & 0.3767 & 0.5160 & 0.3971 & 0.6205 \\
\hspace{1em}MIPROv2 & {\footnotesize\citep{opsahlong2024mipro}} & 0.4589 & 0.5814 & 0.7018 & 0.4510 & 0.3725 & 0.5131 & 0.3987 & 0.6141 \\
\noalign{\jrsavepos\jrposBd}%
\midrule
\rowcolor{gray!12}
\multicolumn{10}{l}{\textsc{Reference lines}} \\
\noalign{\jrsavepos\jrposTe}%
\addlinespace[1pt]
\hspace{1em}\emph{random} & {\footnotesize\emph{mean over ten bge-m3 candidates}} & \emph{0.2798} & \emph{0.2536} & \emph{0.5549} & \emph{0.2507} & \emph{0.2966} & \emph{0.3271} & \emph{0.2351} & \emph{0.3923} \\
\hspace{1em}\emph{rank1} & {\footnotesize\emph{bge-m3 top-1 candidate}} & \emph{0.4211} & \emph{0.5245} & \emph{0.6513} & \emph{0.4384} & \emph{0.3491} & \emph{0.4769} & \emph{0.3595} & \emph{0.5749} \\
\noalign{\global\let\jrhairline\jrhairlinedraw}%
\hspace{1em}\emph{oracle} & {\footnotesize\emph{best of ten, in hindsight}} & \emph{0.7080} & \emph{0.8220} & \emph{0.8438} & \emph{0.6660} & \emph{0.6039} & \emph{0.7288} & \emph{0.5880} & \emph{0.8310} \\
\noalign{\jrsavepos\jrposBe}%
\bottomrule
\end{tabular}
\end{adjustbox}
\par\vskip 9pt\relax
\begin{minipage}{\textwidth}\footnotesize\raggedright
$^{\P}$Min-K\%++ \citep{zhang2025minkpp} is the successor of Min-K\% Prob \citep{shi2024mink}, printed for completeness and not among the fourteen ranked configurations. Our reimplementation matches the official release term for term, and it still scores below \texttt{random} (Appendix~\ref{app:baselines}).
\end{minipage}
\end{table}
Every baseline is re-implemented as a passage selector under the \emph{strongest}
configuration its published settings or released code supports, and evaluated on the same
pools with the same frozen respondent and the same answering template, with $\polar$ the
only additional content \ours{} places in it.
 Fourteen
ranked configurations fall into three families, Table~\ref{tab:main}'s three baseline
group headers, differing in where the selection signal
comes from: (i)~a measurement from models as published, most often of uncertainty;
(ii)~a selector model we train; or (iii)~a polarizer string we search for.
We also report two system-level comparisons in Appendix~\ref{sec:ctrlrag}
(Table~\ref{tab:trainedgen}): Self-RAG's own trained-respondent configuration and our
reproduction of the concurrent CTRL-RAG preprint \citep{ctrlrag2026}. Both train the
answering model itself, so they compare inference classes rather than selection quality
and are not ranked.
Table~\ref{tab:main}'s three reference lines (\texttt{random},
\texttt{rank1}, \texttt{oracle}) calibrate rather than compete.
Per-row configurations,
reference-line definitions, the load-bearing
disclosures about how rows should be read, and every residual deviation from
the official scripts are in Appendix~\ref{app:baselines}.

\paragraph{Published signals re-purposed as selectors.}
Semantic entropy, Self-RAG (reflection), SeaKR, CLeHe, EPR, Min-K\% Prob, IGP,
EigenScore and SPS fill nine rows, each run exactly as published, with no training stage
in our pipeline. Seven of them score each candidate from
the frozen respondent's own response, SPS scores the passage representation alone, and
Self-RAG (reflection) selects with the method's officially released critic, which is what
keeps it under the frozen respondent.

\paragraph{Search-based guidance.}
We also ask whether the polarizer can simply be searched for. GEPA and
MIPROv2 answer that under \ours{}'s exact reward function, training draws and frozen
respondent, so only the optimizer differs.
Both configurations start from the same hand-written caution seed and select identically at
inference ($\arg\min_p H_1(q,p,\psi)$). Each row reports its strongest seed against
\ours{}'s three-seed mean, a comparison deliberately set in the baselines' favour.

\paragraph{Trained rerankers.}
GainRAG, MBA-RAG and CRITIC-R1 replace the signal with a model trained on the same
$1{,}939$ NQ-Open questions as \ours{}, so NQ is in-domain and the other four
out-of-domain.
Each scores the same ten candidates: GainRAG a
$278$M cross-encoder distilled from a gold-conditioned gain teacher, MBA-RAG a
DistilBERT bandit over candidate ranks, and CRITIC-R1 a GRPO-trained critic
reading each cached draft answer. Re-run from stored artifacts on different hardware, all
three reproduce their tabulated values ($\Delta F_1{=}0.0000$ on WebQuestions).

\begin{figure}[t]
\centering
\includegraphics[width=\textwidth]{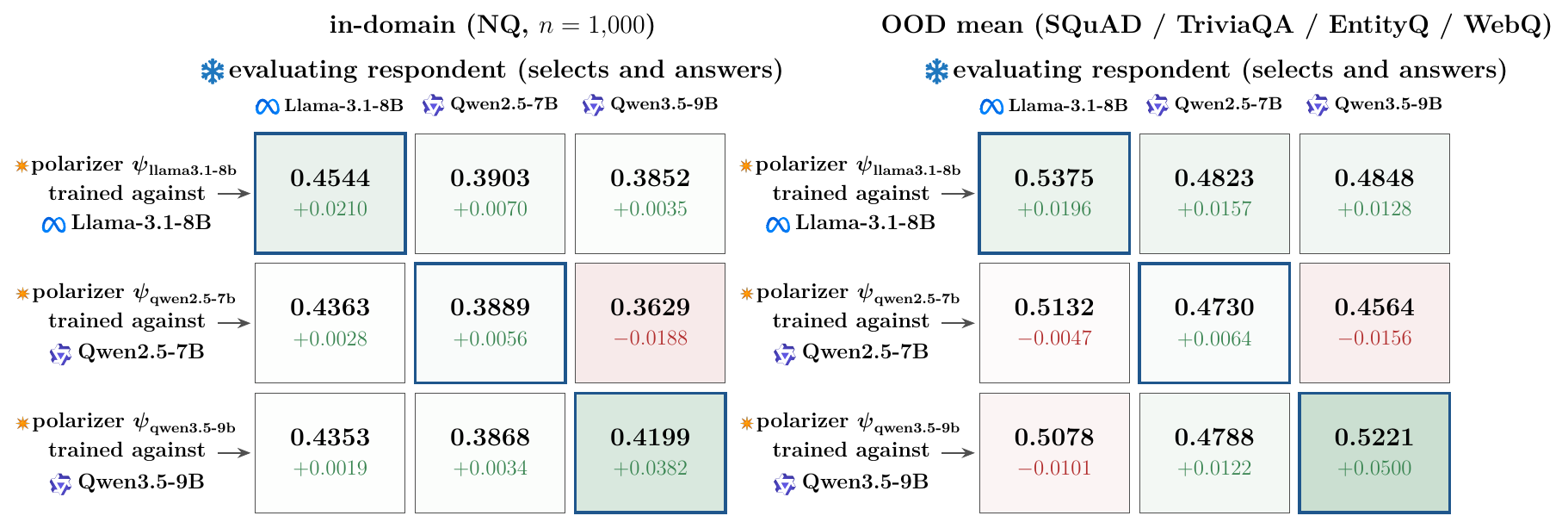}
\caption{\textbf{Cross-respondent transfer of the polarizer under a mean-normalized-entropy
($\bar{H}_{L}$) selector.} Rows are the respondent the polarizer $\psi$ was trained against,
columns the frozen respondent that selects and answers; each of the eighteen facets prints answer $F_1$ with $\psi$ inserted (top)
and $\Delta$ against the same-pass configuration without it (bottom).
Every diagonal is positive on both panels; all four negative facets are off-diagonal. Cell colour
gives the sign of $\Delta$, its depth the magnitude; largest gain $+0.0500$, largest
harm $-0.0188$.}
\label{fig:crossresp-matrix}
\end{figure}

\begin{table}[t]
\centering
\caption{\textbf{The polarizer ablation under two metrics.} Both blocks compare \ours{} against
\ours{} with $\polar$ removed, i.e., plain first-token entropy selection.
Rates are $P(\text{\textsc{misleading}}\mid\text{selected})$ at $K{=}10$ on the bge-m3 pool,
whose own rate is $28.9$; \textsc{macro} averages the five.}
\label{tab:ablation}
\small
\setlength{\tabcolsep}{4pt}
\begin{adjustbox}{max width=\textwidth}
\newlength{\ablcolwd}\settowidth{\ablcolwd}{TriviaQA}%
\newlength{\ablnumwd}\settowidth{\ablnumwd}{$0.4643$}%
\ifdim\ablnumwd>\ablcolwd \setlength{\ablcolwd}{\ablnumwd}\fi
\begin{tabular}{lw{c}{\ablcolwd}w{c}{\ablcolwd}w{c}{\ablcolwd}w{c}{\ablcolwd}w{c}{\ablcolwd}w{c}{\ablcolwd}}
\toprule
Selection rule & NQ & SQuAD & TriviaQA & EntityQ & WebQ & \textsc{macro} \\
\midrule
\rowcolor{gray!12}
\multicolumn{7}{l}{\textsc{Judged-misleading rate of the selected passage} (\%) $\downarrow$} \\
\ours{} w/o polarizer & $36.7$ & $24.7$ & $15.9$ & $39.1$ & $35.0$ & $30.3$ \\
\lodestaricon~\ourstab{} & $\mathbf{32.7}$ & $\mathbf{21.4}$ & $\mathbf{13.9}$ & $\mathbf{32.7}$ & $\mathbf{29.5}$ & $\mathbf{26.0}$ \\
\midrule
\rowcolor{gray!12}
\multicolumn{7}{l}{\textsc{Answer} $F_1$ $\uparrow$} \\
\ours{} w/o polarizer & $0.4643$ & $0.5657$ & $0.7129$ & $0.4610$ & $0.3699$ & $0.5148$ \\
\lodestaricon~\ourstab{} & $\mathbf{0.4789}$ & $\mathbf{0.5968}$ & $\mathbf{0.7181}$ & $\mathbf{0.4839}$ & $\mathbf{0.3917}$ & $\mathbf{0.5339}$ \\
\bottomrule
\end{tabular}
\end{adjustbox}
\end{table}

\section{Main Results}
\label{sec:main}

\paragraph{Head-to-head against published signals.}
\ours{} attains the highest mean $F_1$ of any frozen-respondent method, $0.5339$
against $0.5148$ for plain first-token entropy selection and $0.4769$ for the
retriever's own top-1 passage. It takes the highest exact match ($0.4136$,
Table~\ref{tab:em}) on the same pools, and the highest GPT-4o judge score of the frozen-respondent
configurations judged ($0.6435$, Appendix~\ref{app:judge}).
Throughout this section \emph{mean $F_1$} is the macro mean over the five pools, the
\emph{Mean} column of Table~\ref{tab:main}. The step from the entropy selector holds
on both sides of the split that mean conceals: $0.4643$ to $0.4789$ in-domain on NQ,
and $0.5274$ to $0.5476$ averaged over SQuAD, TriviaQA, EntityQuestions and
WebQuestions. The \ours{} row of Table~\ref{tab:main}, the mean of its three training
seeds, also wins all $70$ method-by-dataset cells on $F_1$, and takes the highest macro
exact match of any ranked row, $0.4136$ against semantic entropy's $0.4039$
(Table~\ref{tab:em}). The sweep is a property of that row rather than of
each run behind it: seed $43$ loses the EntityQuestions $F_1$ cell to GainRAG
($0.4661$ against $0.4675$), while every run does win all five datasets against both
entropy selection and \texttt{rank1} (Appendix~\ref{app:stability}). %
Those cells span the fourteen frozen-respondent-comparable
baseline configurations we rank, meaning every configuration that picks one of those ten
candidates and leaves the respondent frozen.
The three runs beneath the \ours{} row differ in the training-data seed alone. It
reshuffles the grouping and ordering of the $1{,}939$ training questions, while which
questions exist, the train/held-out split and the trainer's own seed are held fixed, so
the three converged polarizers are scored on one held-out set
(Appendix~\ref{app:stability}). Every $F_1$ advantage of \ours{} over a baseline configuration is
paired-significant under both a $t$-test and Wilcoxon (Appendix~\ref{app:significance}).

\paragraph{Reporting conventions.}
Table~\ref{tab:main} reports the mean of three training runs and lists them individually;
the three differ only in their training-data seed, each a $300$-step run over the same
$1{,}939$ questions in a different grouping and order, with their spread reported in
Appendix~\ref{app:stability}. A table that instead reports a single run says so in its caption.
Every headline number uses
the reward of Equation~\ref{eq:reward}.

\section{The Polarizer Ablation}
\label{sec:ablation}

\paragraph{The ablation.}
We ablate \ours{} by removing the polarizer $\polar$ and changing nothing else. The frozen
respondent, the pools and the $\arg\min$ selection rule are identical, so the two configurations differ
only in whether the string is inserted. That configuration is also the strongest untrained selector in
our benchmark (Appendix~\ref{app:baselines}). Table~\ref{tab:ablation} gives it per dataset under two metrics: answer $F_1$, the outcome, and
how often the selected passage is judged \textsc{misleading}, the quantity the reward acts on.
On the latter the string helps on $5/5$ datasets, from $2.0$ points (TriviaQA) to $6.4$
(EntityQuestions), and carries the rule below the pool's own $28.9\%$, which entropy selection
alone does not reach. Averaged over the three training seeds
(Appendix~\ref{app:stability}) the gain over the ablated configuration is $+0.0191 \pm 0.0054$, with
per-dataset gains from $+0.005$ (TriviaQA) to $+0.031$ (SQuAD) and $5/5$ datasets positive
under every seed. That lead is paired-significant under \emph{every} seed individually, the
three seeds giving $p_w = 2.4\times10^{-8}$, $1.8\times10^{-4}$ and
$2.9\times10^{-11}$, under the protocol of Appendix~\ref{app:significance}, whose
table carries the seed-$42$ run alone. Appendix~\ref{app:polarizertext} decomposes the entropy shift the
string produces, by passage class.

\paragraph{Cross-respondent transfer.}
Figure~\ref{fig:crossresp-matrix} repeats the comparison between inserting the polarizer $\polar$
and leaving it out, under a second selection signal (i.e., $\bar{H}_{L}$) and across three
frozen respondents from two model families, Llama-3.1-8B, Qwen2.5-7B and Qwen3.5-9B. All four negative facets are transfers to a respondent the string was
never trained against. In the Qwen3.5-9B column the answer sequence ends at
\texttt{<|endoftext|>} rather than \texttt{<|im\_end|>}, because that is where its generation
actually stops. Restricting it to questions where both answers finished leaves the signs
unchanged. Pooled across the figure's two panels
($5{,}008$ questions) the diagonal stays at $+0.0395$ against $+0.0477$ unrestricted, and the
Llama-3.1-8B transfer grows rather than shrinks.

\section{Conclusion}

We first show that the frozen respondent's first-token entropy chooses
passages well enough to beat the retriever's own ranking with no gold answer, $0.5148$
mean answer $F_1$ against its $0.4769$, but that it fails in a way invisible
from where prior work measures entropy.
The standard entropy selector keeps the lowest-entropy candidate (Equation~\ref{eq:igp})
and is drawn toward the passages it should avoid: pooled over all questions,
judged-misleading candidates carry lower entropy than the rest ($1.22$ nats against $1.33$;
Section~\ref{sec:related}).
Across the five pools it reads a judged-misleading passage more often on average than picking one of
the ten at random (Table~\ref{tab:ablation}).
We propose \lodestarinlineicon\ours{}, which repairs this without touching
the respondent: a GRPO-trained policy writes the polarizer $\polar$, one short
natural-language string whose only job is to navigate the respondent's entropy in the
intended direction.
Across $5{,}008$ questions from five QA benchmarks that one string is enough to give \ours{} the highest mean $F_1$ of any
inference-ready selector, $0.5339$, its three-seed mean winning every method-by-dataset
$F_1$ cell against fourteen published configurations
(Table~\ref{tab:main}). Of that lead, the ablation removing $\polar$
with the respondent, pools and selection rule unchanged attributes
$+0.0191 \pm 0.0054$ $F_1$ to the string itself, positive on $5/5$ datasets under all
three training seeds (Section~\ref{sec:ablation}). All of it
costs one forward pass per candidate and no gold answers. Overall, entropy is the right
signal here but only once directed rather than passively measured: one
learned string turns within-question entropy from a signal that misleads into one that
selects, with the respondent left frozen.

\clearpage
\bibliographystyle{iclr2026_conference}
\bibliography{references}

\newpage
\appendix

\section*{Appendix contents}

\newcommand{\apxline}[2]{%
  \par\noindent
  \makebox[2.4em][l]{\textbf{\ref{#1}}}%
  \parbox[t]{\dimexpr\linewidth-2.4em\relax}{#2}%
  \vspace{2.5pt}%
}

\apxline{app:notation}{\textbf{Notation.} Every symbol used in the paper, collected.}
\apxline{app:grpo}{\textbf{GRPO objective and hyper-parameters.} The objective of
  Equation~\ref{eq:grpo} in full, and every hyper-parameter.}
\apxline{app:alignedprobe}{\textbf{Aligned-probe ablation.} What IGP's gain term scores once
  its probe reads the answering prompt, against first-token entropy on the same five pools.}
\apxline{app:significance}{\textbf{Paired significance tests.} \ours{} against each method on
  $F_1$.}
\apxline{app:stability}{\textbf{Run and seed stability.} What separates the three training
  runs, and where the tested run sits among them.}
\apxline{app:em}{\textbf{Exact-match results and paired tests.} The exact-match counterpart of
  Table~\ref{tab:main}, and the same paired tests on per-question EM differences.}
\apxline{app:judge}{\textbf{LLM-judge scores per dataset.} Table~\ref{tab:main}'s judge column
  opened up pool by pool, including the one pool where the judge and $F_1$ disagree.}
\apxline{app:polarizertext}{\textbf{Learned polarizer text and its effect on the entropy.} The
  converged strings themselves, the two framings independent seeds reach under a reward that
  never scores wording, and how the string shifts the entropy of supporting and
  non-supporting passages.}
\apxline{sec:ctrlrag}{\textbf{CTRL-RAG reproduction.} Its strongest checkpoint, why no
  single checkpoint of that run can stand in for the method, what its reward was maximising
  while the format decayed, and the scope limits on what we do and do not claim at its
  published scale.}
\apxline{app:baselines}{\textbf{Baseline configurations and disclosures.} Official settings, the
  disclosures about how individual rows should be read, and every residual deviation from
  the published scripts.}

\clearpage
\section{Notation}
\label{app:notation}
Table~\ref{tab:notation} collects the paper's symbols for reference.

\begin{table}[h]
\centering
\caption{\textbf{Symbols used in the paper.} Defined at first use in
Sections~\ref{sec:method} and~\ref{sec:settings}; collected here for reference.}
\label{tab:notation}
\setlength{\tabcolsep}{4pt}
\small
\begin{tabular}{l p{0.70\textwidth}}
\toprule
Symbol & Meaning \\
\midrule
$q$ & a question \\
$p$ & a candidate passage \\
$P(q) = \{p_1,\dots,p_K\}$ & the retrieved candidate set for $q$; we retrieve with bge-m3 \citep{chen2024bgem3} \\
$K$ & the candidate-pool depth $|P(q)|$; $K{=}10$ throughout the main results \\
$R$ & the frozen respondent, marked \frozenmark{} where the tables record it: the LLM that answers from $q$ and one $p$, parameters never updated (Llama-3.1-8B-Instruct) \\
$H_1(q,p)$ & Shannon entropy of $R$'s first answer-token distribution, no polarizer \\
$H_1(q,p,\psi)$ & the same entropy with polarizer $\psi$ inserted \\
$\bar{H}_{L}(q,p)$ & mean normalized entropy of the $L$ answer tokens $R$ generates from
  passage $p$ alone, no polarizer \\
$\bar{H}_{L}(q,p,\psi)$ & the same entropy with polarizer $\psi$ inserted \\
$\mathrm{Mis}(q),\ \mathrm{Sup}(q)$ & the misleading / supporting passages of $q$ \\
$\Delta_{\mathrm{Mis}-\mathrm{Sup}} \bar{H}_{L}(q)$ & within-question separation, Equation~\ref{eq:sep} \\
$\Delta_{\mathrm{Mis}-\mathrm{Sup}} \bar{H}_{L}(q;\psi)$ & within-question separation with $\psi$ inserted, Equation~\ref{eq:reward} \\
$\psi$ & a sampled polarizer; $\psi_i$ the $i$-th of a group \\
$\polar$ & the \emph{inference-time} polarizer: the single converged output. The star marks
  what survives to inference, so the policy $\pi_\theta$ carries none \\
$\pi_\theta,\ \pi_{\text{ref}}$ & the GRPO polarizer policy and its frozen initialization \\
$\mathcal{S}(\psi)$ & reward of polarizer $\psi$: the \emph{separation} its insertion
  induces, baseline-corrected, Equation~\ref{eq:reward} \\
$G$ & group size (number of polarizers sampled per prompt) \\
$\hat{A}_i$ & group-standardized advantage of $\psi_i$, Equation~\ref{eq:adv} \\
$\rho_{i,t}(\theta)$ & token-level importance ratio \\
$a(q,p)$ & the answer $R$ generates from passage $p$ alone \\
$a(q,p;\psi)$ & the same answer with polarizer $\psi$ inserted; Equation~\ref{eq:select} keeps the entropy-minimizing one \\
$\hat{a}$ & the selected answer, Equation~\ref{eq:select} \\
\bottomrule
\end{tabular}
\end{table}

\section{GRPO objective and hyper-parameters}
\label{app:grpo}

This section states the GRPO objective in full:
\begin{equation}
\mathcal{J}(\theta) \;=\; \mathbb{E}\Big[
\min\!\big(\rho\,\hat{A},\;
\operatorname{clip}(\rho,\,1-\varepsilon_{\text{low}},\,1+\varepsilon_{\text{high}})\,\hat{A}\big)
\Big] \;-\; \beta\, \mathbb{D}_{\mathrm{KL}}\big[\pi_\theta \,\Vert\, \pi_{\text{ref}}\big],
\label{eq:grpo}
\end{equation}
where $\rho$ is the token-level importance ratio against $\pi_{\theta_{\text{old}}}$,
$\rho_{i,t}(\theta) = \pi_\theta(\psi_{i,t} \mid x, \psi_{i,<t}) \,/\,
\pi_{\theta_{\text{old}}}(\psi_{i,t} \mid x, \psi_{i,<t})$, and $\pi_{\text{ref}}$ is the
frozen initial policy. At each training step $\pi_\theta$ samples a group of $G$
polarizers from the single fixed prompt $x$, and all $G$ candidates are scored by
Equation~\ref{eq:reward} on the same batch of labelled questions; a step draws $16$
such groups, the $128$ sampled strings of Table~\ref{tab:grpo-hparams}. GRPO
\citep{shao2024deepseekmath} replaces a learned value baseline
with the statistics of the sampled group: the advantage of polarizer $\psi_i$ is its reward
standardized over the $G$ rewards of that group,
\begin{equation}
\hat{A}_i \;=\; \frac{\mathcal{S}(\psi_i) \;-\; \operatorname{mean}\big\{\mathcal{S}(\psi_j)\big\}_{j=1}^{G}}
                     {\operatorname{std}\big\{\mathcal{S}(\psi_j)\big\}_{j=1}^{G}},
\label{eq:adv}
\end{equation}
broadcast to every token of $\psi_i$. Following \citet{liu2025drgrpo} and
\citet{yu2025dapo}, we aggregate token-mean over the group, so the sequence-level reward
is not silently re-weighted by polarizer length. The clip range is asymmetric,
$(\varepsilon_{\text{low}}, \varepsilon_{\text{high}}) = (0.2, 0.28)$, and the KL term
uses $\beta = 10^{-3}$ with the low-variance $k_3$ estimator, kept in the
loss rather than in the reward so that Equation~\ref{eq:reward} remains the \emph{only}
quantity the advantage ranks. Because all $G$ polarizers of a group are scored on the
same questions under the same frozen respondent, this advantage ranks nothing but
phrasings of the polarizer.

Table~\ref{tab:grpo-hparams} lists the configuration behind
Equations~\ref{eq:grpo} and~\ref{eq:adv}. Every polarizer reported in this paper is read at
the final step of the $300$-step budget, with \emph{no} per-run checkpoint selection of
any kind.
The polarizer policy emits its polarizer wrapped in
\texttt{<critique>...</critique>} tags, a name inherited from the training harness whose
contents are the polarizer $\psi$ of Section~\ref{sec:method} and nothing else, and the
string is parsed out before scoring. The policy prompt contains only the fixed
category-level instruction and unlabelled exemplar question--passage pairs, never a
label, a gold answer, or respondent feedback. $\polar$ itself is inserted into the
respondent's prompt rather than the policy's, after the passage $p$ and before the
question $q$: it reads \texttt{Passages:}~$p$ / \texttt{An expert's analysis of
the passage above:}~$\polar$ / \texttt{Question:}~$q$.

\begin{table}[h]
\centering
\caption{\textbf{GRPO configuration} used for every training run reported in this
paper.}
\label{tab:grpo-hparams}
\setlength{\tabcolsep}{4pt}
\small
\begin{tabular}{ll}
\toprule
Setting & Value \\
\midrule
Polarizer policy $\pi_\theta$ & Qwen3-4B-Instruct \\
Reference policy $\pi_{\text{ref}}$ & frozen initial policy \\
Group size $G$ & $8$ \\
Question groups per step & $16$ ($128$ sampled strings per step) \\
Training budget & $300$ steps (one epoch of the $4{,}800$-row file from the $1{,}939$-question pool) \\
Sampling temperature & $1.1$ \\
Maximum polarizer length & $96$ tokens \\
Advantage & group-standardized (Equation~\ref{eq:adv}) \\
Loss aggregation & token-mean over the group \\
Clip range $(\varepsilon_{\text{low}}, \varepsilon_{\text{high}})$ & $(0.2,\,0.28)$; dual-clip constant $3.0$ \\
KL regularization & in-loss, low-variance $k_3$ estimator, $\beta = 10^{-3}$ \\
Reward & Equation~\ref{eq:reward}; per-passage entropy shifts clipped to $\pm 2$ \\
Malformed samples & fixed penalty in place of a respondent evaluation \\
\midrule
Trainer hardware & $4\times$ NVIDIA RTX PRO 6000 Blackwell ($96$\,GB) \\
Wall-clock & $\approx 7$\,h for the $300$-step run (median $85$\,s/step) \\
\bottomrule
\end{tabular}
\end{table}

\section{Aligned-probe ablation}
\label{app:alignedprobe}
\label{par:igp-aligned}
Appendix~\ref{app:baselines}'s IGP entry attributes part of its position to a prompt mismatch rather than to its
gain formulation: its published implementation probes with its own short prompt, whereas
the polarizer ablation's $H_1(q,p)$ (Section~\ref{sec:ablation}) is read from the very
prompt that produces the scored answer. We isolate that factor by re-scoring IGP's
per-passage all-token mean entropy $\bar{H}_{L}(q,p)$ under the shared answering prompt,
with the same frozen respondent on the same five pools
(Table~\ref{tab:igp-aligned}). The two macro means are close, $0.5148$ for $H_1$ against
$0.5001$ for the aligned all-token mean, but that closeness holds only on average: per
pool the two disagree in both directions, $H_1$ ahead by $+0.0339$ on TriviaQA and the
aligned mean ahead by $0.0070$ on SQuAD.

\begin{table}[h]
\centering
\caption{\textbf{First-token entropy against the same-prompt all-token mean that IGP's gain
term uses}, on the five canonical pools, scored with the frozen respondent under the shared
answering prompt (as in Table~\ref{tab:main}). The two macro means are close ($0.5148$ vs.\
$0.5001$), but per pool the two variants disagree in both directions: $H_1$ leads by its
largest margin on TriviaQA ($+0.0339$), while the aligned all-token mean leads by its largest
margin on SQuAD ($-0.0070$); the small mean gap is not per-pool equivalence.}
\label{tab:igp-aligned}
\setlength{\tabcolsep}{4pt}
\small
\begin{adjustbox}{max width=\textwidth}
\begin{tabular}{lw{c}{\fonecolwd}w{c}{\fonecolwd}w{c}{\fonecolwd}w{c}{\fonecolwd}w{c}{\fonecolwd}w{c}{\fonecolwd}}
\toprule
 & \multicolumn{6}{c}{\textbf{Answer $F_1$} $\uparrow$ (generated vs.\ gold)} \\
\cmidrule(lr){2-7}
\multirow{2}{*}{Variant}
  & In-domain & \multicolumn{4}{c}{Out-of-domain} & \multirow{2}{*}{Mean} \\
\cmidrule(lr){2-2} \cmidrule(lr){3-6}
  & NQ & SQuAD & TriviaQA & EntityQ & WebQ & \\
\midrule
$H_1(q,p)$, first-token entropy & 0.4643 & 0.5657 & 0.7129 & 0.4610 & 0.3699 & 0.5148 \\
$\bar{H}_{L}(q,p)$, aligned all-token mean & 0.4364 & 0.5727 & 0.6791 & 0.4402 & 0.3722 & 0.5001 \\
\midrule
$\Delta$ ($H_1 - \bar{H}_{L}$) & $+0.0279$ & $-0.0070$ & $+0.0339$ & $+0.0208$ & $-0.0023$ & $+0.0146$ \\
\bottomrule
\end{tabular}
\end{adjustbox}
\end{table}

The narrowing does not move IGP's ranked row. Every baseline is ranked under the strongest
configuration its published settings or released code supports
(Section~\ref{sec:settings-baselines}). Probing with the answering prompt is a change we make
to IGP's gain term, not an option those settings offer. Table~\ref{tab:main} therefore keeps
IGP at its published configuration's macro mean of $0.4844$, and $0.5001$ stands as an
ablation result. The aligned probe recovers about half the distance from that row to $H_1$'s
$0.5148$, and leaves $H_1$ ahead on the macro mean.

\section{Paired significance tests}
\label{app:significance}

Because all methods select from the same ten candidates of the same question, we test
per-question $F_1$ differences with paired bootstrap ($10{,}000$ resamples), a paired
$t$-test and Wilcoxon signed-rank (Table~\ref{tab:sig}), every row of it scored on all
five pools and all $5{,}008$ questions. Every test in this appendix is
paired on the seed-$42$ run's per-question selections, that being the fixed-budget
run whose polarizer is used at inference; Table~\ref{tab:main}'s \ours{} row is
instead the mean of three training runs, and is characterized by its $\pm 0.0054$
spread rather than by a $p$-value. Three rows carry a disclosure with their marker. The
$^{\S\S}$ configuration's original run predates the per-question logging these tests need and was
re-run with it, greedy CLeHe reproducing its tabulated macro $F_1$ to $-0.0007$, so the
test is of the re-run. The $^{\ast}$ configuration is reconstructed under first-wins tie-breaking,
which \emph{favours} official semantic entropy: its tie rate is $62$--$79\%$ with
$\sim\!2$ distinct scores per $10$ candidates, so first-wins imports a rank-1 prior.
Removing it (\texttt{random\_tiebreak}) puts that method at mean $F_1$ $0.4774$,
indistinguishable from \texttt{rank1}'s $0.4769$. The $^{\parallel}$ row's paired
vector comes from EPR's official-budget configuration (Appendix~\ref{app:baselines}),
macro $0.4955$, rather than from the ranked $24$-token row, macro $0.4968$, so its
$\Delta F_1$ reads $+0.0389$ where the ranked row's macro difference is $+0.0375$.

\begin{table}[!tp]
\centering
\caption{\textbf{Paired significance of \ours{} against each method} on per-question
$F_1$ differences. Every row is tested against the same \ours{} run on the same
questions, so the rows differ only in the method being compared. $\Delta F_1$ is that
difference in answer $F_1$ as defined in Section~\ref{sec:settings}, reported with its
$95\%$ bootstrap interval and two paired tests, and bold marks $p<0.05$ on both. Rows are
banded by method family, in Table~\ref{tab:main}'s order; $^{\S\S}$ marks a configuration re-run
for per-question logging and $^{\ast}$ a configuration reconstructed under first-wins
tie-breaking, and $^{\parallel}$ one whose paired vector is its official-budget run; all
three are stated in full in the text.}
\label{tab:sig}
\setlength{\tabcolsep}{3.5pt}
\small
\begin{adjustbox}{max width=\textwidth}
\begin{tabular}{llrlrr}
\toprule
Method & Reference & $\Delta F_1$ & $95\%$ CI & $p_t$ & $p_{\text{Wilcoxon}}$ \\
\midrule
\rowcolor{gray!12}
\multicolumn{6}{l}{\textsc{Published signals re-purposed as selectors}} \\
\addlinespace[1pt]
\hspace{1em}Semantic entropy$^{\ast}$ & {\footnotesize\citep{farquhar2024semantic}} & $\mathbf{+0.0246}$ & $[+0.0149,\, +0.0342]$ & $4.0{\times}10^{-7}$ & $4.4{\times}10^{-7}$ \\
\hspace{1em}Self-RAG (reflection) & {\footnotesize\citep{asai2024selfrag}} & $\mathbf{+0.0272}$ & $[+0.0171,\, +0.0373]$ & $1.8{\times}10^{-7}$ & $4.1{\times}10^{-7}$ \\
\hspace{1em}SeaKR & {\footnotesize\citep{yao2025seakr}} & $\mathbf{+0.0295}$ & $[+0.0207,\, +0.0383]$ & $7.3{\times}10^{-11}$ & $7.9{\times}10^{-11}$ \\
\hspace{1em}CLeHe$^{\S\S}$ & {\footnotesize\citep{clehe2025}} & $\mathbf{+0.0348}$ & $[+0.0258,\, +0.0438]$ & $2.3{\times}10^{-14}$ & $5.1{\times}10^{-14}$ \\
\hspace{1em}EPR$^{\parallel}$ & {\footnotesize\citep{epr2026}} & $\mathbf{+0.0389}$ & $[+0.0299,\, +0.0480]$ & $3.5{\times}10^{-17}$ & $1.1{\times}10^{-16}$ \\
\hspace{1em}Min-K\% Prob & {\footnotesize\citep{shi2024mink}} & $\mathbf{+0.0420}$ & $[+0.0330,\, +0.0512]$ & $1.3{\times}10^{-19}$ & $4.1{\times}10^{-19}$ \\
\hspace{1em}IGP & {\footnotesize\citep{song2026igp}} & $\mathbf{+0.0500}$ & $[+0.0398,\, +0.0602]$ & $2.6{\times}10^{-21}$ & $1.2{\times}10^{-20}$ \\
\hspace{1em}EigenScore & {\footnotesize\citep{chen2024inside}} & $\mathbf{+0.0526}$ & $[+0.0427,\, +0.0622]$ & $5.1{\times}10^{-26}$ & $4.4{\times}10^{-25}$ \\
\hspace{1em}SPS & {\footnotesize\citep{sps2026}} & $\mathbf{+0.1965}$ & $[+0.1840,\, +0.2091]$ & $8.4{\times}10^{-206}$ & $3.8{\times}10^{-169}$ \\
\midrule
\rowcolor{gray!12}
\multicolumn{6}{l}{\textsc{Trained reranker (respondent-free)}} \\
\addlinespace[1pt]
\hspace{1em}GainRAG & {\footnotesize\citep{gainrag2025}} & $\mathbf{+0.0393}$ & $[+0.0289,\, +0.0496]$ & $6.3{\times}10^{-14}$ & $2.2{\times}10^{-13}$ \\
\hspace{1em}MBA-RAG & {\footnotesize\citep{tang2025mbarag}} & $\mathbf{+0.0579}$ & $[+0.0468,\, +0.0689]$ & $4.3{\times}10^{-25}$ & $4.4{\times}10^{-24}$ \\
\hspace{1em}CRITIC-R1 & {\footnotesize\citep{criticr12026}} & $\mathbf{+0.0719}$ & $[+0.0613,\, +0.0826]$ & $4.6{\times}10^{-39}$ & $1.3{\times}10^{-36}$ \\
\midrule
\rowcolor{gray!12}
\multicolumn{6}{l}{\textsc{Search-based guidance}} \\
\addlinespace[1pt]
\hspace{1em}GEPA & {\footnotesize\citep{agrawal2025gepa}} & $\mathbf{+0.0183}$ & $[+0.0105,\, +0.0262]$ & $4.1{\times}10^{-6}$ & $5.7{\times}10^{-6}$ \\
\hspace{1em}MIPROv2 & {\footnotesize\citep{opsahlong2024mipro}} & $\mathbf{+0.0212}$ & $[+0.0134,\, +0.0290]$ & $1.4{\times}10^{-7}$ & $1.8{\times}10^{-7}$ \\
\midrule
\rowcolor{gray!12}
\multicolumn{6}{l}{\textsc{Reference lines}} \\
\addlinespace[1pt]
\hspace{1em}\emph{rank1} &  & $\mathbf{+0.0575}$ & $[+0.0464,\, +0.0685]$ & $9.1{\times}10^{-25}$ & $9.0{\times}10^{-24}$ \\
\hspace{1em}\emph{oracle} &  & \emph{$-0.1943$} & \emph{$[-0.2035,\, -0.1853]$} & $\approx 0$ & $5.6{\times}10^{-287}$ \\
\midrule
\rowcolor{gray!12}
\multicolumn{6}{l}{\textsc{Polarizer ablation}} \\
\addlinespace[1pt]
\hspace{1em}\ours{} w/o polarizer &  & $\mathbf{+0.0197}$ & $[+0.0128,\, +0.0267]$ & $2.1{\times}10^{-8}$ & $2.4{\times}10^{-8}$ \\
\bottomrule
\end{tabular}
\end{adjustbox}
\end{table}

Every $\Delta F_1$, $95\%$ confidence interval and $p$-value behind these comparisons is
in Table~\ref{tab:sig}, and the exact-match tests are in the last two columns of
Table~\ref{tab:em} (Appendix~\ref{app:em}).
``Winning all $70$ cells'' is a statement about cell-wise
values rather than about per-dataset statistical significance; the statistically
grounded form of the claim is ``highest mean $F_1$, paired-significant on both tests
against all fourteen baseline configurations ranked''.

\paragraph{Plain entropy against the retrieval prior, respondent by respondent.}
The polarizer's value is easiest to read where the signal it steers has none. On NQ, and
within a single pass that recomputes both configurations question by question ($n = 1{,}000$),
selection by first-token $H_1$ alone is statistically indistinguishable from taking the
retriever's top-ranked passage on the two Qwen respondents. Each figure
comes from that respondent's own pass, which recomputes both configurations together, so the
contrast is exactly paired within it. On Qwen2.5-7B the difference is $+0.0139$
$[-0.0113,\, +0.0391]$, $p_w = 0.42$; on Qwen3.5-9B it is $+0.003$, $p_w = 0.64$.
Llama-3.1-8B is the one respondent of the three on which the untrained signal separates
from the retrieval prior by itself\textemdash{}$0.4643$ against \texttt{rank1}'s $0.4211$
on its own NQ pass, and $0.5148$ against $0.4769$ over the five pools
(Table~\ref{tab:main}). Each
respondent's polarizer is trained on that same first-token statistic, so the comparison is
matched, and inserting it carries those two past \texttt{rank1}, to $0.3701$ over $0.3452$
and $0.4086$ over $0.3517$. Two qualifications belong beside that. The all-token estimator, which we do not
train on, \emph{does} separate from \texttt{rank1} on Qwen2.5-7B ($+0.0395$
$[+0.0143,\, +0.0649]$, $p_w = 2.8\times10^{-3}$), so the failure is first-token $H_1$'s
rather than every untrained signal's. And a signal's standalone value does not predict what
the reward can extract from it, Qwen3.5-9B having the weakest standalone signal of the three
and the largest gain over its own $H_1$ configuration, $+0.0544$, $p_w = 8.3\times10^{-10}$.

\section{Run and seed stability}
\label{app:stability}

Table~\ref{tab:main} reports \ours{} as the mean of three training runs and lists those
runs beneath its \ours{} row. This appendix says what separates them, which is the
\emph{training-data seed} and nothing else. Each of the three is its own $300$-step run
under the inference-time configuration, evaluated at its final checkpoint, and each
consumes its own training file built by the same builder under seed $42$, $43$ or $44$.
That seed draws $4{,}800$ rows of $16$ question groups per step from the same
$1{,}939$-question training pool, so $300$ steps is one epoch over the file and what the
seed varies is the grouping and ordering of the training questions, not which questions
exist. Nothing else differs: the train/held-out split seed is pinned at $42$ in all
three, so they share one held-out split byte for byte, and the trainer's own seed is
never overridden. They land at mean $F_1$ $0.5343$ (seed $42$), $0.5283$ (seed $43$) and
$0.5390$ (seed $44$), that is $0.5339 \pm 0.0054$, with a gain over entropy of
$+0.0191 \pm 0.0054$ and every run winning all five datasets against both entropy
selection and \texttt{rank1}. Two properties of that spread matter for how the rest of
the paper should be read. No run involves a checkpoint choice of any kind, each being
the final step of the fixed budget (Appendix~\ref{app:grpo}). And the seed-$42$ run,
which carries every paired test in this paper, is the \emph{median} of the three rather
than the best of them, so the tested run is a representative draw and not a selected
peak.

\paragraph{An execution floor for that spread.}
A fourth run places the $\pm 0.0054$ against something. The seed-$42$ configuration was
trained a second time on the identical training file, again for the full $300$-step budget
and read at its own final checkpoint, and it lands at $0.5333$ against that seed's
$0.5343$. What separates the two is execution alone, meaning floating-point accumulation
order, rollout sampling, and how the reward-scoring workers were distributed across
devices. The variability
therefore has three levels of its own. A same-seed re-run moves the five-dataset mean by
$0.0010$, a change of training-data seed moves it by $0.0054$ (sd), and the method effect
across the three seeds runs from $+0.0135$ to $+0.0243$, so the quantity this paper
reports is more than an order of magnitude above the noisiest level beneath it.

\section{Exact-match results and paired tests}
\label{app:em}

Table~\ref{tab:em} repeats the comparison of Table~\ref{tab:main} under exact
match, with the same three training seeds averaged into the \ours{} row and
listed beneath it. The ordering is largely preserved and every claim made on
$F_1$ holds under EM. Against the fourteen ranked configurations \ours{} takes the
highest macro exact match, $0.4136$ against official semantic entropy's $0.4039$, and it
does so without winning every cell: on WebQuestions the three-seed mean trails that method,
$0.2283$ against $0.2300$, a deficit belonging to the mean more than to every run, since
seed $44$ takes the cell on its own at $0.2310$ while seeds $42$ and $43$ land at $0.2280$
and $0.2260$. The system-level Self-RAG row,
marked $^{\S}$ in the table, is excluded from that count, because it answers with its own
trained generator in an extraction configuration and is not
frozen-respondent-comparable; that disclosure is given under \emph{Trained generators} in
Appendix~\ref{app:baselines}, with every other per-row configuration. The
\textsc{polarizer ablation} row carries no seed breakdown, that configuration having no polarizer
and being identical under all three training seeds. Our CTRL-RAG reproduction is not a
row here either, as in Table~\ref{tab:main}; it is reported in the case study of
Appendix~\ref{sec:ctrlrag}, where its exact-match numbers are tabulated.

The last two columns of Table~\ref{tab:em} repeat the paired tests of
Table~\ref{tab:sig} on the per-question EM differences of the same $5{,}008$ aligned
questions. Their values are aligned on the $\times$ so that the exponents can be compared
down the column, and every one is printed rather than marked, so how close a comparison
runs to $0.05$ can be read off its own row.
Every conclusion of the $F_1$ tests survives under EM, the narrowest margin being the
lead over official semantic entropy, which clears $0.05$ on both tests but by less than
any other row ($p_t = 2.3\times10^{-2}$,
$p_{\text{Wilcoxon}} = 4.8\times10^{-2}$), which is unsurprising for a coarser
integer-valued score on what was already the closest comparison under $F_1$. The polarizer ablation, the cleanest test of the contribution,
remains significant under EM on both tests.

\paragraph{LLM-as-judge protocol.}
Because both $F_1$ and EM are lexical, we additionally audit answer correctness
with an LLM judge, so that no conclusion rests on string overlap alone. To keep
the audit independent of our own design choices, the judging prompt is adopted
\emph{verbatim} from the official code release of the semantic-entropy paper
\citep{farquhar2024semantic}, in its multi-reference branch, since every dataset here
ships a list of gold answers. The judge is
\texttt{gpt-4o-2024-08-06} \citep{openai2024gpt4o}, queried at temperature $0$ with a ten-token decoding
budget and constrained to a yes/no verdict, which we map to $1$ and $0$;
a reply that is neither is re-queried once and otherwise counted
as incorrect. With the per-example fields as placeholders, the prompt is reproduced in
Table~\ref{tab:judgeprompt}.

\begin{table}[!b]
\centering
\caption{\textbf{The LLM-judge prompt}, adopted \emph{verbatim} from the official
release of \citet{farquhar2024semantic} in its multi-reference branch. Braced tokens
are the per-example fields substituted at query time.}
\label{tab:judgeprompt}
\begin{promptbox}[Prompt template]
{\bf Instruction}\\
\small\ttfamily\raggedright
We are assessing the quality of answers to the following question:
\{question\}\\
The following are expected answers to this question: \{correct\_answers\}.\\
The proposed answer is: \{predicted\_answer\}\\
Within the context of the question, does the proposed answer mean the same as
any of the expected answers? Respond only with yes or no.
\tcblower
{\bf Constrained answer cue}\\
\small\ttfamily\raggedright
Response:
\end{promptbox}
\end{table}

\makeatletter
\settowidth{\sfp@cell}{\ttfamily\scriptsize M}
\settowidth{\sfp@colwd}{\small{\footnotesize\citep{opsahlong2024mipro}}}
\begingroup\small\global\sfp@rowht=\dimexpr4\baselineskip\relax\endgroup
\sfp@build
\makeatother
\begin{table}[!tp]
\centering
\caption{\textbf{Exact match on the same pools and methods as Table~\ref{tab:main}.}
Its rows differ from that table's in two places: Min-K\%++, printed there for
completeness, is not repeated, and the system-level Self-RAG row ($^{\S}$) is added.
Every cell is $1$ where the respondent's generated answer matches a gold answer after
SQuAD normalization and $0$ otherwise, averaged over the pool, so the rows carry that
table's one-row-per-method rule and family grouping under a second metric.
The last two columns are that row's paired tests against \ours{} on per-question EM
differences, under the protocol of Appendix~\ref{app:significance}; a \cna{} there marks
a row that admits no per-question vector, and $^{\S}$ a system-level row, whose test
therefore compares inference classes rather than selection quality.
The \ours{} row is the mean of the three training seeds listed beneath it, as in
Table~\ref{tab:main}; reference lines keep their band's italic, and the \emph{oracle}
pair is the one comparison that runs against \ours{} rather than for it.}
\label{tab:em}
\setlength{\tabcolsep}{3.5pt}
\small
\begin{adjustbox}{max width=\textwidth}
\newlength{\pflank}\newlength{\ptesthd}\newlength{\ptimeswd}%
\begingroup\small
\settowidth{\pflank}{10\textsuperscript{\textminus 147}}%
\settowidth{\ptimeswd}{\,\texttimes\,}%
\settowidth{\dimen2}{\cna}%
\ifdim\ptimeswd<\dimen2 \setlength{\ptimeswd}{\dimen2}\fi
\settowidth{\ptesthd}{\textbf{Paired tests vs.\ \ourstab{}}}%
\global\pflank=\pflank \global\ptimeswd=\ptimeswd \global\ptesthd=\ptesthd
\endgroup
\newlength{\pouter}\newlength{\pinner}%
\setlength{\pouter}{\dimexpr0.25\ptesthd-0.5\ptimeswd\relax}%
\ifdim\pouter<\pflank \setlength{\pouter}{\pflank}\fi
\setlength{\pinner}{\dimexpr\pouter-\tabcolsep\relax}%
\ifdim\pinner<\pflank \setlength{\pinner}{\pflank}\fi
\newlength{\plabl}\newlength{\plabr}\newlength{\pshift}\newlength{\pfirst}\newlength{\plast}%
\begingroup\small
\settowidth{\plabl}{$p_t$}%
\settowidth{\plabr}{$p_{\text{Wilcoxon}}$}%
\global\plabl=\plabl \global\plabr=\plabr
\endgroup
\setlength{\pshift}{\dimexpr0.25\plabr-0.25\plabl\relax}%
\setlength{\pfirst}{\dimexpr\pouter-\pshift\relax}%
\setlength{\plast}{\dimexpr\pouter+\pshift\relax}%
\begin{tabular}{llw{c}{\fonecolwd}w{c}{\fonecolwd}w{c}{\fonecolwd}w{c}{\fonecolwd}w{c}{\fonecolwd}w{c}{\fonecolwd}w{r}{\pfirst}@{}c@{}w{l}{\pinner} w{r}{\pinner}@{}c@{}w{l}{\plast}}
\toprule
 & & \multicolumn{6}{c}{\textbf{Answer exact match (EM)} $\uparrow$}
   & \multicolumn{6}{c}{\textbf{Paired tests vs.\ \ourstab{}}} \\
\cmidrule(lr){3-8} \cmidrule(lr){9-14}
\multirow{2}{*}{Method} & \multirow{2}{*}{Reference}
  & In-domain & \multicolumn{4}{c}{Out-of-domain} & \multirow{2}{*}{Mean}
  & & \multirow{2}{*}{\makebox[0pt][c]{$p_t$}} &
  & & \multirow{2}{*}{\makebox[0pt][c]{$p_{\text{Wilcoxon}}$}} & \\
\cmidrule(lr){3-3} \cmidrule(lr){4-7}
  & & NQ & SQuAD & TriviaQA & EntityQ & WebQ & & & & & & & \\
\midrule
\rowcolor{gray!12}
\multicolumn{14}{l}{\textsc{RL-learned polarizer}} \\
\addlinespace[1pt]
\hspace{1em}\textbf{\lodestaricon~\ourstab{}} & \multirow{4}{*}{\sfpfit} & \textbf{0.3663} & \textbf{0.4817} & \textbf{0.6423} & \textbf{0.3492} & \textbf{0.2283} & \textbf{0.4136} & & \cna & & & \cna & \\
\hspace{2em} -- seed 42 & & 0.3670 & 0.4770 & 0.6420 & 0.3601 & 0.2280 & 0.4148 & & \cna & & & \cna & \\
\hspace{2em} -- seed 43 & & 0.3620 & 0.4800 & 0.6470 & 0.3383 & 0.2260 & 0.4107 & & \cna & & & \cna & \\
\hspace{2em} -- seed 44 & & 0.3700 & 0.4880 & 0.6380 & 0.3492 & 0.2310 & 0.4152 & & \cna & & & \cna & \\
\midrule
\rowcolor{gray!12}
\multicolumn{14}{l}{\textsc{Published signals re-purposed as selectors}} \\
\addlinespace[1pt]
\hspace{1em}Semantic entropy & {\footnotesize\citep{farquhar2024semantic}} & 0.3590 & 0.4710 & 0.6300 & 0.3294 & 0.2300 & 0.4039 & 2.3 & \,\texttimes\, & 10\textsuperscript{\textminus 2} & 4.8 & \,\texttimes\, & 10\textsuperscript{\textminus 2} \\
\hspace{1em}SeaKR & {\footnotesize\citep{yao2025seakr}} & 0.3450 & 0.4630 & 0.6210 & 0.3274 & 0.2230 & 0.3959 & 4.4 & \,\texttimes\, & 10\textsuperscript{\textminus 5} & 4.1 & \,\texttimes\, & 10\textsuperscript{\textminus 4} \\
\hspace{1em}Self-RAG (reflection) & {\footnotesize\citep{asai2024selfrag}} & 0.3450 & 0.4450 & 0.6070 & 0.3135 & 0.2060 & 0.3833 & 4.3 & \,\texttimes\, & 10\textsuperscript{\textminus 9} & 4.0 & \,\texttimes\, & 10\textsuperscript{\textminus 7} \\
\hspace{1em}Min-K\% Prob & {\footnotesize\citep{shi2024mink}} & 0.3260 & 0.4410 & 0.6150 & 0.3026 & 0.2140 & 0.3797 & 2.1 & \,\texttimes\, & 10\textsuperscript{\textminus 13} & 2.5 & \,\texttimes\, & 10\textsuperscript{\textminus 10} \\
\hspace{1em}CLeHe & {\footnotesize\citep{clehe2025}} & 0.3210 & 0.4530 & 0.5960 & 0.3095 & 0.2120 & 0.3783 & 2.3 & \,\texttimes\, & 10\textsuperscript{\textminus 15} & 9.0 & \,\texttimes\, & 10\textsuperscript{\textminus 12} \\
\hspace{1em}EPR & {\footnotesize\citep{epr2026}} & 0.3210 & 0.4440 & 0.5890 & 0.3075 & 0.2150 & 0.3753 & 6.1 & \,\texttimes\, & 10\textsuperscript{\textminus 18} & 1.1 & \,\texttimes\, & 10\textsuperscript{\textminus 13} \\
\hspace{1em}EigenScore & {\footnotesize\citep{chen2024inside}} & 0.3120 & 0.4430 & 0.5930 & 0.3026 & 0.2050 & 0.3711 & 1.6 & \,\texttimes\, & 10\textsuperscript{\textminus 17} & 2.4 & \,\texttimes\, & 10\textsuperscript{\textminus 13} \\
\hspace{1em}IGP & {\footnotesize\citep{song2026igp}} & 0.3150 & 0.4050 & 0.6030 & 0.2986 & 0.1820 & 0.3607 & 3.6 & \,\texttimes\, & 10\textsuperscript{\textminus 23} & 1.8 & \,\texttimes\, & 10\textsuperscript{\textminus 17} \\
\hspace{1em}SPS & {\footnotesize\citep{sps2026}} & 0.2120 & 0.1800 & 0.4970 & 0.1657 & 0.1650 & 0.2439 & 7.1 & \,\texttimes\, & 10\textsuperscript{\textminus 147} & 6.0 & \,\texttimes\, & 10\textsuperscript{\textminus 98} \\
\midrule
\rowcolor{gray!12}
\multicolumn{14}{l}{\textsc{Trained reranker (respondent-free)}} \\
\addlinespace[1pt]
\hspace{1em}GainRAG & {\footnotesize\citep{gainrag2025}} & 0.3280 & 0.4010 & 0.6040 & 0.3194 & 0.2010 & 0.3707 & 9.0 & \,\texttimes\, & 10\textsuperscript{\textminus 16} & 4.5 & \,\texttimes\, & 10\textsuperscript{\textminus 12} \\
\hspace{1em}MBA-RAG & {\footnotesize\citep{tang2025mbarag}} & 0.3140 & 0.4120 & 0.5750 & 0.3056 & 0.1890 & 0.3591 & 7.4 & \,\texttimes\, & 10\textsuperscript{\textminus 22} & 1.6 & \,\texttimes\, & 10\textsuperscript{\textminus 16} \\
\hspace{1em}CRITIC-R1 & {\footnotesize\citep{criticr12026}} & 0.3010 & 0.3800 & 0.5480 & 0.2669 & 0.1930 & 0.3378 & 2.6 & \,\texttimes\, & 10\textsuperscript{\textminus 42} & 4.3 & \,\texttimes\, & 10\textsuperscript{\textminus 31} \\
\midrule
\rowcolor{gray!12}
\multicolumn{14}{l}{\textsc{Search-based guidance}} \\
\addlinespace[1pt]
\hspace{1em}MIPROv2 & {\footnotesize\citep{opsahlong2024mipro}} & 0.3460 & 0.4740 & 0.6220 & 0.3323 & 0.2190 & 0.3987 & 1.2 & \,\texttimes\, & 10\textsuperscript{\textminus 4} & 8.8 & \,\texttimes\, & 10\textsuperscript{\textminus 4} \\
\hspace{1em}GEPA & {\footnotesize\citep{agrawal2025gepa}} & 0.3430 & 0.4680 & 0.6290 & 0.3274 & 0.2180 & 0.3971 & 1.7 & \,\texttimes\, & 10\textsuperscript{\textminus 5} & 2.0 & \,\texttimes\, & 10\textsuperscript{\textminus 4} \\
\midrule
\rowcolor{gray!12}
\multicolumn{14}{l}{\textsc{Trained generator}} \\
\addlinespace[1pt]
\hspace{1em}Self-RAG$^{\S}$ & {\footnotesize\citep{asai2024selfrag}} & 0.3990 & 0.0800 & 0.2990 & 0.0923 & 0.1120 & 0.1965 & 1.2 & \,\texttimes\, & 10\textsuperscript{\textminus 201} & 1.3 & \,\texttimes\, & 10\textsuperscript{\textminus 128} \\
\midrule
\rowcolor{gray!12}
\multicolumn{14}{l}{\textsc{Reference lines}} \\
\addlinespace[1pt]
\hspace{1em}\emph{random} & & \emph{0.1928} & \emph{0.1731} & \emph{0.4795} & \emph{0.1752} & \emph{0.1550} & \emph{0.2351} & & \cna & & & \cna & \\
\hspace{1em}\emph{rank1} & & \emph{0.3160} & \emph{0.4120} & \emph{0.5750} & \emph{0.3056} & \emph{0.1890} & \emph{0.3595} & \emph{1.5} & \emph{\,\texttimes\,} & \emph{10\textsuperscript{\textminus 21}} & \emph{2.6} & \emph{\,\texttimes\,} & \emph{10\textsuperscript{\textminus 16}} \\
\hspace{1em}\emph{oracle} & & \emph{0.5710} & \emph{0.6930} & \emph{0.7700} & \emph{0.5099} & \emph{0.3960} & \emph{0.5880} & \emph{2.4} & \emph{\,\texttimes\,} & \emph{10\textsuperscript{\textminus 229}} & \emph{3.5} & \emph{\,\texttimes\,} & \emph{10\textsuperscript{\textminus 143}} \\
\midrule
\rowcolor{gray!12}
\multicolumn{14}{l}{\textsc{Polarizer ablation}} \\
\addlinespace[1pt]
\hspace{1em}\ours{} w/o polarizer & {\footnotesize\emph{$\polar$ removed: plain $\arg\min_p H_1$}} & 0.3620 & 0.4600 & 0.6400 & 0.3433 & 0.2160 & 0.4043 & 2.1 & \,\texttimes\, & 10\textsuperscript{\textminus 3} & 7.6 & \,\texttimes\, & 10\textsuperscript{\textminus 3} \\
\bottomrule
\end{tabular}
\end{adjustbox}
\end{table}

\section{LLM-judge scores per dataset}
\label{app:judge}

Table~\ref{tab:main} reports the LLM-judge column as a five-pool mean; Table~\ref{tab:judge}
opens it up per dataset, in the same layout and row order. The $F_1$ ordering is reproduced
pool by pool with one exception worth stating: on WebQuestions the Self-RAG (reflection) configuration
leads the seed-$42$ run that carries this table's paired tests, $0.5510$ against $0.5430$
($0.5497$ for the three-seed mean), where under token $F_1$ \ours{} wins that cell. The
disagreement is confined to one pool and does not move the mean, where \ours{} leads the
best baseline by $+0.0208$ on that run and by $+0.0223$ on the three-seed mean, but it is
the kind of per-pool reversal a five-pool average hides, which is why we print the
breakdown.

\makeatletter
\settowidth{\sfp@cell}{\ttfamily\scriptsize M}
\settowidth{\sfp@colwd}{\small{\footnotesize\citep{opsahlong2024mipro}}}
\begingroup\small\global\sfp@rowht=\dimexpr4\baselineskip\relax\endgroup
\sfp@build
\makeatother
\begin{table}[!tp]
\centering
\caption{\textbf{LLM-judge accuracy per dataset} (GPT-4o answer-equivalence, five pools),
opening up the five-pool mean of Table~\ref{tab:main}. Same rows and grouping as
Table~\ref{tab:em}, with \ours{} in bold and reference lines in italic. The last two columns are that
row's paired tests against \ours{} on per-question judge verdicts, under the protocol of
Appendix~\ref{app:significance}, and the \emph{oracle} pair is the one comparison that
runs against \ours{} rather than for it. The \ours{} row is the mean of the three training
seeds listed beneath it, as in Table~\ref{tab:main}, while the two test columns read the
seed-$42$ run alone. A \cna{} is never a zero and never a test that fell short: in the
first six columns it is a measurement this table does not report, and in the last two a
comparison we did not run.}
\label{tab:judge}
\setlength{\tabcolsep}{4pt}
\small
\begin{adjustbox}{max width=\textwidth}
\newlength{\jpflank}\newlength{\jptesthd}\newlength{\jptimeswd}
\begingroup\small
\settowidth{\jpflank}{10\textsuperscript{\textminus 255}}%
\settowidth{\jptimeswd}{\,\texttimes\,}%
\settowidth{\dimen2}{\cna}%
\ifdim\jptimeswd<\dimen2 \setlength{\jptimeswd}{\dimen2}\fi
\settowidth{\jptesthd}{\textbf{Paired tests vs.\ \ourstab{}}}%
\global\jpflank=\jpflank \global\jptimeswd=\jptimeswd \global\jptesthd=\jptesthd
\endgroup
\newlength{\jpouter}\newlength{\jpinner}
\setlength{\jpouter}{\dimexpr0.25\jptesthd-0.5\jptimeswd\relax}
\ifdim\jpouter<\jpflank \setlength{\jpouter}{\jpflank}\fi
\setlength{\jpinner}{\dimexpr\jpouter-\tabcolsep\relax}
\ifdim\jpinner<\jpflank \setlength{\jpinner}{\jpflank}\fi
\begin{tabular}{llw{c}{\fonecolwd}w{c}{\fonecolwd}w{c}{\fonecolwd}w{c}{\fonecolwd}w{c}{\fonecolwd}w{c}{\fonecolwd}w{r}{\jpouter}@{}c@{}w{l}{\jpinner} w{r}{\jpinner}@{}c@{}w{l}{\jpouter}}
\toprule
 & & In-domain & \multicolumn{4}{c}{Out-of-domain} & & \multicolumn{6}{c}{\textbf{Paired tests vs.\ \ourstab{}}} \\
\cmidrule(lr){3-3} \cmidrule(lr){4-7} \cmidrule(lr){9-14}
Method & Reference & NQ & SQuAD & TriviaQA & EntityQ & WebQ & Mean & & \makebox[0pt][c]{$p_t$} & & & \makebox[0pt][c]{$p_{\text{Wilcoxon}}$} & \\
\midrule
\rowcolor{gray!12}
\multicolumn{14}{l}{\textsc{RL-learned polarizer}} \\
\addlinespace[1pt]
\hspace{1em}\textbf{\lodestaricon~\ourstab{}} & \multirow{4}{*}{\sfpfit} & \textbf{0.5803} & \textbf{0.7237} & \textbf{0.7570} & \textbf{0.6068} & \textbf{0.5497} & \textbf{0.6435} & & \cna & & & \cna & \\
\hspace{2em} -- seed 42 & & 0.5780 & 0.7210 & 0.7530 & 0.6151 & 0.5430 & 0.6420 & & \cna & & & \cna & \\
\hspace{2em} -- seed 43 & & 0.5720 & 0.7220 & 0.7590 & 0.5833 & 0.5460 & 0.6365 & & \cna & & & \cna & \\
\hspace{2em} -- seed 44 & & 0.5910 & 0.7280 & 0.7590 & 0.6220 & 0.5600 & 0.6520 & & \cna & & & \cna & \\
\midrule
\rowcolor{gray!12}
\multicolumn{14}{l}{\textsc{Published signals re-purposed as selectors}} \\
\addlinespace[1pt]
\hspace{1em}Self-RAG (reflection) & {\footnotesize\citep{asai2024selfrag}} & 0.5520 & 0.6870 & 0.7220 & 0.5942 & 0.5510 & 0.6212 & 7.6 & \,\texttimes\, & 10\textsuperscript{\textminus 4} & 3.6 & \,\texttimes\, & 10\textsuperscript{\textminus 3} \\
\hspace{1em}Semantic entropy & {\footnotesize\citep{farquhar2024semantic}} & 0.5520 & 0.6610 & 0.7390 & 0.5317 & 0.5170 & 0.6001 & 3.5 & \,\texttimes\, & 10\textsuperscript{\textminus 13} & 3.6 & \,\texttimes\, & 10\textsuperscript{\textminus 10} \\
\hspace{1em}CLeHe & {\footnotesize\citep{clehe2025}} & 0.5540 & 0.6840 & 0.7320 & 0.5377 & 0.5240 & 0.6063 & 5.1 & \,\texttimes\, & 10\textsuperscript{\textminus 11} & 1.5 & \,\texttimes\, & 10\textsuperscript{\textminus 8} \\
\hspace{1em}EPR & {\footnotesize\citep{epr2026}} & 0.5530 & 0.6740 & 0.7270 & 0.5357 & 0.5230 & 0.6025 & 8.0 & \,\texttimes\, & 10\textsuperscript{\textminus 13} & 6.7 & \,\texttimes\, & 10\textsuperscript{\textminus 10} \\
\hspace{1em}SeaKR & {\footnotesize\citep{yao2025seakr}} & 0.5390 & 0.6860 & 0.7330 & 0.5218 & 0.5230 & 0.6006 & 1.9 & \,\texttimes\, & 10\textsuperscript{\textminus 14} & 4.2 & \,\texttimes\, & 10\textsuperscript{\textminus 11} \\
\hspace{1em}IGP & {\footnotesize\citep{song2026igp}} & 0.5450 & 0.6170 & 0.7290 & 0.5526 & 0.5240 & 0.5935 & 1.9 & \,\texttimes\, & 10\textsuperscript{\textminus 15} & 7.9 & \,\texttimes\, & 10\textsuperscript{\textminus 12} \\
\hspace{1em}Min-K\% Prob & {\footnotesize\citep{shi2024mink}} & 0.5350 & 0.6470 & 0.7280 & 0.5179 & 0.5150 & 0.5886 & 1.3 & \,\texttimes\, & 10\textsuperscript{\textminus 21} & 2.4 & \,\texttimes\, & 10\textsuperscript{\textminus 16} \\
\hspace{1em}EigenScore & {\footnotesize\citep{chen2024inside}} & 0.5050 & 0.6600 & 0.7090 & 0.5060 & 0.4970 & 0.5754 & 8.6 & \,\texttimes\, & 10\textsuperscript{\textminus 30} & 3.1 & \,\texttimes\, & 10\textsuperscript{\textminus 22} \\
\hspace{1em}SPS & {\footnotesize\citep{sps2026}} & 0.3690 & 0.3260 & 0.5930 & 0.2986 & 0.4510 & 0.4075 & 5.0 & \,\texttimes\, & 10\textsuperscript{\textminus 212} & 3.9 & \,\texttimes\, & 10\textsuperscript{\textminus 134} \\
\midrule
\rowcolor{gray!12}
\multicolumn{14}{l}{\textsc{Trained reranker (respondent-free)}} \\
\addlinespace[1pt]
\hspace{1em}GainRAG & {\footnotesize\citep{gainrag2025}} & 0.5560 & 0.6120 & 0.7240 & 0.6002 & 0.5350 & 0.6054 & 3.9 & \,\texttimes\, & 10\textsuperscript{\textminus 9} & 3.7 & \,\texttimes\, & 10\textsuperscript{\textminus 7} \\
\hspace{1em}MBA-RAG & {\footnotesize\citep{tang2025mbarag}} & 0.5120 & 0.6220 & 0.6840 & 0.5625 & 0.4930 & 0.5747 & 3.3 & \,\texttimes\, & 10\textsuperscript{\textminus 24} & 3.2 & \,\texttimes\, & 10\textsuperscript{\textminus 18} \\
\hspace{1em}CRITIC-R1 & {\footnotesize\citep{criticr12026}} & 0.5290 & 0.6130 & 0.6870 & 0.5040 & 0.5230 & 0.5712 & 1.6 & \,\texttimes\, & 10\textsuperscript{\textminus 27} & 1.3 & \,\texttimes\, & 10\textsuperscript{\textminus 20} \\
\midrule
\rowcolor{gray!12}
\multicolumn{14}{l}{\textsc{Search-based guidance}} \\
\addlinespace[1pt]
\hspace{1em}GEPA & {\footnotesize\citep{agrawal2025gepa}} & 0.5670 & 0.7100 & 0.7490 & 0.5496 & 0.5270 & 0.6205 & 7.0 & \,\texttimes\, & 10\textsuperscript{\textminus 6} & 1.0 & \,\texttimes\, & 10\textsuperscript{\textminus 4} \\
\hspace{1em}MIPROv2 & {\footnotesize\citep{opsahlong2024mipro}} & 0.5530 & 0.6900 & 0.7490 & 0.5635 & 0.5150 & 0.6141 & 9.3 & \,\texttimes\, & 10\textsuperscript{\textminus 9} & 7.0 & \,\texttimes\, & 10\textsuperscript{\textminus 7} \\
\midrule
\rowcolor{gray!12}
\multicolumn{14}{l}{\textsc{Trained generator}} \\
\addlinespace[1pt]
\hspace{1em}Self-RAG$^{\S}$ & {\footnotesize\citep{asai2024selfrag}} & 0.5190 & 0.6110 & 0.6850 & 0.6161 & 0.5570 & 0.5976 & 2.6 & \,\texttimes\, & 10\textsuperscript{\textminus 10} & 5.0 & \,\texttimes\, & 10\textsuperscript{\textminus 8} \\
\midrule
\rowcolor{gray!12}
\multicolumn{14}{l}{\textsc{Reference lines}} \\
\addlinespace[1pt]
\hspace{1em}\emph{random} & & \emph{0.3560} & \emph{0.3195} & \emph{0.5797} & \emph{0.3008} & \emph{0.4057} & \emph{0.3923} & & \cna & & & \cna & \\
\hspace{1em}\emph{rank1} & & \emph{0.5130} & \emph{0.6220} & \emph{0.6840} & \emph{0.5625} & \emph{0.4930} & \emph{0.5749} & \emph{4.4} & \emph{\,\texttimes\,} & \emph{10\textsuperscript{\textminus 24}} & \emph{3.9} & \emph{\,\texttimes\,} & \emph{10\textsuperscript{\textminus 18}} \\
\hspace{1em}\emph{oracle} & & \emph{0.8240} & \emph{0.9220} & \emph{0.8710} & \emph{0.7778} & \emph{0.7600} & \emph{0.8310} & \emph{1.4} & \emph{\,\texttimes\,} & \emph{10\textsuperscript{\textminus 255}} & \emph{2.1} & \emph{\,\texttimes\,} & \emph{10\textsuperscript{\textminus 156}} \\
\midrule
\rowcolor{gray!12}
\multicolumn{14}{l}{\textsc{Polarizer ablation}} \\
\addlinespace[1pt]
\hspace{1em}\ours{} w/o polarizer & {\footnotesize\emph{$\polar$ removed: plain $\arg\min_p H_1$}} & 0.5600 & 0.6830 & 0.7490 & 0.5585 & 0.5140 & 0.6129 & 3.4 & \,\texttimes\, & 10\textsuperscript{\textminus 12} & 1.9 & \,\texttimes\, & 10\textsuperscript{\textminus 9} \\
\bottomrule
\end{tabular}
\end{adjustbox}
\end{table}

Every mean in Table~\ref{tab:judge} reproduces the corresponding cell of
Table~\ref{tab:main} exactly wherever both tables report one, which is the check that the
two columns are the same measurement and not two independent judge runs. Semantic entropy's
$0.6001$ is opened up per pool here as well, having been obtained the same way: the corrected
run's cached selections looked up in that one judging pass, with no query of its own.

The two paired-test columns are the one place in this table where the three training seeds
do not average. Their \ours{} configuration is the seed-$42$ run's per-question verdict file, whose
five means are the seed-$42$ row, so neither test speaks for the three-seed mean that the
\ours{} row and Table~\ref{tab:main} report. They add no query to the judging the rest of
the table already rests on, that pass having scored all ten candidates of every one of the
$5{,}008$ questions, so each configuration's per-question verdict is the verdict at the rank it
selected. A \cna{} in those two columns is therefore a comparison we did not run rather
than a test that ran and fell short, and only two kinds of row still carry one. \ours{}
and its seeds are the comparison's own reference, and the \emph{random} line is an
expectation over the ten ranks rather than a per-question selection, so neither admits a
paired vector at all. The system-level Self-RAG row is tested and reads $^{\S}$: that
method trains its own answering model, so its test compares inference classes rather than
selection quality and the row stays outside the fourteen frozen-respondent-comparable
configurations of Table~\ref{tab:main}, which it is not a row of. The values themselves
are aligned on the $\times$ so that the exponents can be compared down the column.

\paragraph{Auditing the labelling judge.}
The judge above scores answers; the passage labels of Section~\ref{sec:method} come from
a separate pair of labelling judges, whose inter-judge agreement is Cohen's $\kappa = 0.675$. A separate
benchmark audits the same judge pair against an independent 70B bf16 third-party judge,
Llama-3.1-70B-Instruct \citep{grattafiori2024llama3}, on $1{,}300$ held-out rows: agreement
with the dual-judge consensus label, measured on the $600$ consensus rows, is $96\%$, with
a $1.5\%$ false-positive rate on $200$ no-injection anchors. The two figures are not the
same measurement and do not confirm each other: $\kappa$ is chance-corrected and the
$96\%$ is not, and the two are computed on different sets of rows.

\section{Learned polarizer text and its effect on the entropy}
\label{app:polarizertext}

The converged strings are interpretable, and independent seeds agree on what to say
even though the reward never scores wording. The inference-time $\polar$ opens by
\emph{directing doubt}: ``\emph{The passage may address a similar entity or context
but \textbf{not} the specific one asked for, and it might contain outdated, fictional,
or misattributed information}\ldots''. The second seed converges on a paraphrase of the
same instruction: ``\emph{a related or similarly named topic but does \textbf{not}
directly or correctly answer the questioned topic; it could be referencing a different
entity, event, or timeframe}\ldots''. Since the reward is defined purely on $H_1$
separation, this convergence is evidence that the objective has a stable linguistic
optimum rather than that any phrasing was supplied.

The opening clause is not, however, a reliable predictor of quality. The third seed
instead opens by \emph{affirming}: ``\emph{The passage may \textbf{directly address}
the surface-level query but could mislead by providing outdated, incomplete, or
contextually adjacent information}\ldots''. It reaches mean $F_1$ $0.5283$, below both
direct-doubt runs ($0.5343$ and $0.5390$) yet above plain entropy selection's $0.5148$
on the same pools.
All three runs win all five
datasets against both entropy selection and \texttt{rank1}
(Appendix~\ref{app:stability}). Affirm-then-qualify and direct-doubt framings are
therefore both reachable optima of the same objective, and we report the wording as
interpretable rather than as diagnostic.

\begin{table}[!t]
\centering
\caption{\textbf{The polarizer across the $300$-step budget}, greedy-decoded from the five
retained checkpoints of the deployed run, one box per checkpoint. The decode is still moving
through the first three; steps $240$ and $300$ emit byte-identical strings, so the four boxes
cover all five checkpoints, and the last of them is the $\polar$ every number in this paper is
computed from.}
\label{tab:psitraj}
\begin{tcolorbox}
{\bf Step 60 --- still moving}\\
\small\ttfamily\raggedright\setlength{\parindent}{0pt}
The passage may directly address a surface-level fact but could mislead if it
refers to a related entity, a different time period, a variant form of the subject, or a
fictionalized version rather than the actual entity. It might also contain outdated
statistics, approximate numbers, or misattributed details that appear plausible but are not
accurate for the specific query.
\end{tcolorbox}
\vspace{2pt}
\begin{tcolorbox}
{\bf Step 120 --- still moving}\\
\small\ttfamily\raggedright\setlength{\parindent}{0pt}
The passage may address a similar entity or context but not the specific one
asked for, and it might contain outdated, approximate, or misattributed figures that do not
accurately reflect the correct quantity or detail. It could also misrepresent the scope or
boundaries of the subject, leading to a misleading answer.
\end{tcolorbox}
\vspace{2pt}
\begin{tcolorbox}
{\bf Step 180 --- still moving}\\
\small\ttfamily\raggedright\setlength{\parindent}{0pt}
The passage may address a similar entity or context but not the specific one
asked for, and it might contain outdated, approximate, or misattributed figures that
misrepresent the correct answer.
\end{tcolorbox}
\vspace{2pt}
\begin{tcolorbox}
{\bf Steps 240 and 300 --- byte-identical, and the deployed $\polar$}\\
\small\ttfamily\raggedright\setlength{\parindent}{0pt}
The passage may address a similar entity or context but not the specific
one asked for, and it might contain outdated, fictional, or misattributed information that
misrepresents the true answer.
\end{tcolorbox}
\end{table}

\paragraph{How the string is reached, and in what sense it is one string.}
Table~\ref{tab:psitraj} shows $\polar$ at five checkpoints spanning the $300$-step budget of
the deployed run. The wording is not stable early: across the ten checkpoints dumped every
$30$ steps the greedy decode takes seven distinct values, and the early strings are longer
and carry extra qualifying clauses ($375$ characters at step $60$ against $194$ at step
$300$). The deployed string first appears at step $240$ and is byte-identical at $270$ and
$300$.

Within a step, $\polar$ is the survivor of a group rather than a single output: $\pi_\theta$
samples $G=8$ polarizers at temperature $1.1$ and GRPO ranks them against one another
(Table~\ref{tab:grpo-hparams}). Those training rollouts were not written to disk, so the
within-step spread \emph{during} training cannot be recovered from this run. What can be
measured is the converged policy re-sampled afterwards, and it is close to a point mass:
drawing $8$ samples from the final checkpoint at the training temperature returns eight
byte-identical strings, and widening to $512$ draws at temperature $1.0$ returns the deployed
$\polar$ $502$ times ($98.0\%$), the ten that differ occurring once each and moving only
inside the clause that lists what may be wrong with the passage. A control at temperature
$2.0$ returns six distinct strings from eight draws, so the sampler is live rather than
silently greedy. This measures the policy after training rather than the group the optimizer
ranked at step $300$, and it speaks to how the reported string was obtained, not to whether
the polarizer helps.

\paragraph{What the string does to the entropy.}
Decomposing the shift $H_1(q,p,\polar) - H_1(q,p)$ by passage class over the five pools, a
zero-inference calculation on the cached probes, shows $\polar$ acting almost entirely on
one side. Passages the respondent answers correctly from move by $+0.007$ on average and
passages it fails on by $+0.258$, a difference-in-differences of $+0.251$. The effect
concentrates where selection fails. The per-question \emph{lure}, the wrong passage with
the lowest plain entropy and therefore the one a plain-entropy selector actually picks, is
pushed up by $+0.377$, against $+0.106$ for that question's best supporting passage.
These figures are a descriptive decomposition of cached probe values, not a causal
account of what the reward taught the polarizer policy.

\section{CTRL-RAG reproduction}
\label{sec:ctrlrag}

CTRL-RAG \citep{ctrlrag2026} is a concurrent preprint that trains the generator for
context faithfulness rather than a selector, so it is a paradigm relative rather than a
same-task competitor. We reproduced it end-to-end and report it as a case study rather
than as a row of Tables~\ref{tab:main}, \ref{tab:sig}
and~\ref{tab:em} for one measurable reason: \emph{the run has no stable
protocol-compliant checkpoint}, so every candidate number would misrepresent it, and
each would do so differently. The clean short-answer format survives to roughly
$90$--$106$ steps of the $300$-step protocol; after that the raw output collapses and
stays collapsed\textemdash{}macro $F_1$ $0.1344$ at step $120$, $0.1276$ at step $300$
(Table~\ref{tab:ctrlrag}), \emph{every} exact match lost at both\textemdash{}and the
extraction rule we
pre-registered against the earlier failure mode fires on only $9\%$--$10\%$ of
questions against the next variant.
Table~\ref{tab:trainedgen} still prints both trained generators beside \ours{}, and each
of its two rows needs its provenance stated. The CTRL-RAG cells are the strongest
checkpoint of the run, RL step $80$, so Table~\ref{tab:ctrlrag}'s trajectory rather than
that row is the reproduction's full record, and its judge column carries the same five-pool
macro, so step $80$ reads $0.6925$ in both tables.
The Self-RAG cells are that method's system-level row, the
configuration its $^{\S}$ disclosure attaches to in Table~\ref{tab:em}, where its
per-pool exact match is printed. Self-RAG still selects one passage, as every ranked row
does, while CTRL-RAG reads all ten in retrieval order and exposes no selection step at
all.

\begin{table}[!tp]
\centering
\caption{\textbf{Trained generators beside \ours{}, in the format of
Table~\ref{tab:main}.} Both comparators train the answering model itself, so these are
comparisons of inference class rather than of selection quality, and the two rows sit
outside Table~\ref{tab:main}'s frozen-respondent ranking. The \textsc{respondent}
column marks that model as frozen (\cfr) or fine-tuned (\cft), and the judge column is
the five-pool GPT-4o macro of Appendix~\ref{app:judge}'s protocol.}
\label{tab:trainedgen}
\setlength{\tabcolsep}{3.5pt}
\small
\begin{adjustbox}{max width=\textwidth}
\begin{tabular}{llcw{c}{\fonecolwd}w{c}{\fonecolwd}w{c}{\fonecolwd}w{c}{\fonecolwd}w{c}{\fonecolwd}w{c}{\fonecolwd}w{c}{\emjudgecolwd}}
\toprule
 & & \multirow{3}{*}{\textsc{Respondent}}
   & \multicolumn{6}{c}{\textbf{Answer $F_1$} $\uparrow$\;\;(generated answer vs.\ gold answers)} & \\
\cmidrule(lr){4-9}
\multirow{2}{*}{Method} & \multirow{2}{*}{Reference} &
  & In-domain & \multicolumn{4}{c}{Out-of-domain} & \multirow{2}{*}{Mean} & \multirow{2}{*}{\shortstack{LLM-Judge\\(GPT-4o)}} \\
\cmidrule(lr){4-4} \cmidrule(lr){5-8}
  & & & NQ & SQuAD & TriviaQA & EntityQ & WebQ & & \\
\midrule
\hspace{1em}\lodestaricon~\ourstab{} & & \cfr & 0.4783 & 0.5916 & 0.7163 & 0.4943 & 0.3908 & 0.5343 & 0.6420 \\
\midrule
\hspace{1em}Self-RAG & {\footnotesize\citep{asai2024selfrag}} & \cft & 0.4796 & 0.2587 & 0.4684 & 0.2977 & 0.3147 & 0.3638 & 0.5976 \\
\hspace{1em}CTRL-RAG & {\footnotesize\citep{ctrlrag2026}} & \cft & 0.5492 & 0.6318 & 0.7198 & 0.5817 & 0.4775 & 0.5920 & 0.6925 \\
\bottomrule
\end{tabular}
\end{adjustbox}
\end{table}

\paragraph{The strongest checkpoint, stated plainly.}
The reproduction beats \ours{} on every pool, its best checkpoint reaching mean $F_1$
$\mathbf{0.5920}$ against \ours{}'s $0.5343$ (Table~\ref{tab:ctrlrag}). That reference row
is the seed-$42$ training run in every column, where Table~\ref{tab:main} reports the
three-seed mean, $0.5339$ macro $F_1$ and $0.4136$ EM. Every row of the table is read
under the raw protocol, on which the pre-registered extraction step changes step $20$ by
less than $10^{-4}$, and the mid RL row is the last checkpoint whose sampled responses
never hit the length cap. Against step
$20$, on the same $5{,}008$ aligned questions, the paired difference runs
\emph{against} us, $\Delta F_1 = -0.0533$ $[-0.0646, -0.0420]$ and $-0.0665$
$[-0.0792, -0.0538]$ under EM. Two qualifications belong next to that number, and
neither subtracts from it. The comparison is one of inference class rather than
selection quality, since this checkpoint is a gold-supervised, fully fine-tuned copy of
the very respondent every ranked row keeps frozen and it reads all ten passages per
query. What that class costs is the concrete form of the distinction. It requires gold
answer supervision, per-domain fine-tuning, serving of the respondent itself and a
ten-passage prefill per query, none of which is available in the frozen-respondent
settings this paper targets, where \ours{} selects at one forward pass per candidate,
needs no gold answers, and serves no model beyond the respondent that answers.
And the margin is bought by \emph{supervised fine-tuning on in-benchmark gold
answers, not by the reinforcement objective the method is about}: the SFT anchor scores
$0.5860$ before a single RL step, three hundred steps of RL add at most $+0.0060$ over
it, and then remove everything. We report the anchor row precisely so that this
decomposition is visible rather than implied.

\begin{figure}[H]
\centering
\includegraphics[width=\textwidth]{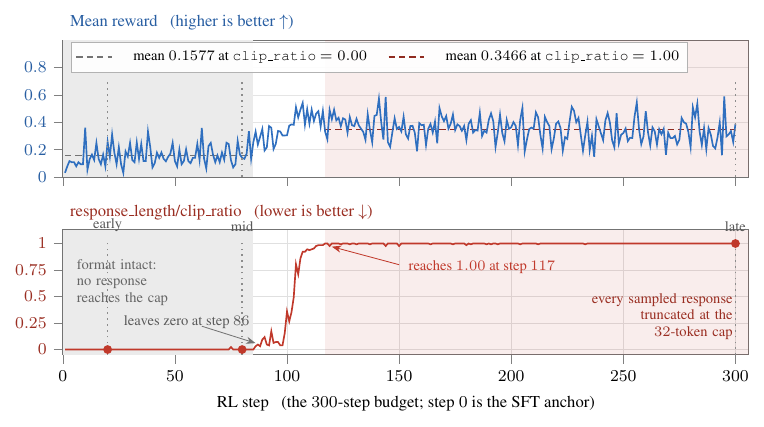}
\caption{\textbf{CTRL-RAG reproduction, the RL stage measured two ways.} Both panels are the
SFT-anchored seed-$42$ run whose checkpoints Table~\ref{tab:ctrlrag} reports, read from its
training log. \emph{Top:} \texttt{critic/score/mean}, the quantity the objective maximises.
\emph{Bottom:} the fraction of sampled responses truncated at the $32$-token cap. That cap is
not binding on a well-formed answer: under the policy's own tokenizer the $1{,}848$ gold answer
strings of the NQ pool average $3.7$ tokens and the longest runs to $13$, so truncation records
a change of output format rather than an answer that did not fit. The reward
rises as the format degrades, from mean $0.1577$ over the $84$ steps at which no response
reaches the cap to mean $0.3466$ over the $165$ steps at which every response does.}
\label{fig:ctrlrag-reward}
\end{figure}

\paragraph{What the judge says about the collapse.}
At step $300$ every string metric is destroyed\textemdash{}$F_1$ $0.1276$ and EM
exactly $0.0000$\textemdash{}yet the GPT-4o judge still credits $0.5845$ of answers
as equivalent, against $0.6727$--$0.6925$ for the three intact checkpoints. Roughly
$85\%$ of the judge score survives an $F_1$ collapse to a fifth of its value, which is
direct evidence that the collapse is a change of surface form and not a loss of content.
Every figure in that column is a five-pool macro, so it reads both down the rows of
Table~\ref{tab:ctrlrag} and against the macros quoted elsewhere, where \ours{} stands at
$0.6435$ over three seeds (Table~\ref{tab:judge}) and at $0.6420$ on the seed-$42$ run
this table's reference row reports.
Figure~\ref{fig:ctrlrag-reward} plots what the run was maximising while that surface form
decayed. The reward rises over exactly the steps in which the sampled responses pin to the
length cap, and CTRL-RAG's objective scores no property of the output format. The surface form
that the string metrics of Table~\ref{tab:ctrlrag} measure is therefore not a quantity this run
was optimising; what it was optimising rewards length outright, as the next paragraph derives.

\begin{table}[H]
\centering
\caption{\textbf{CTRL-RAG reproduction, stage by stage.} The four CTRL-RAG rows are its
SFT anchor at zero RL steps and three points early, mid and late in the $300$-step
budget, all read under the raw protocol, so a column is comparable down the rows.
Those three points are fixed by the run rather than selected from it: $20$ is the first
checkpoint it wrote under its $20$-step save interval, $80$ is the last whose sampled
responses never reach the length cap (Figure~\ref{fig:ctrlrag-reward}; the next saved
checkpoint, $100$, already truncates $36\%$ of them), and $300$ is the final step of the
budget, the checkpoint Appendix~\ref{app:grpo}'s no-per-run-selection rule reports for every
trained configuration.
\emph{Judge} is the GPT-4o answer-equivalence protocol of Table~\ref{tab:main}, macro-averaged
over the same five pools as every other judge figure in the paper. Bold marks the best of the
four CTRL-RAG rows, and \ours{} is a reference row that is the seed-$42$ training run in
every column, where Table~\ref{tab:main} reports the three-seed mean.}
\label{tab:ctrlrag}
\setlength{\tabcolsep}{4pt}
\small
\begin{adjustbox}{max width=\textwidth}
\begin{tabular}{lw{c}{\fonecolwd}w{c}{\fonecolwd}w{c}{\fonecolwd}w{c}{\fonecolwd}w{c}{\fonecolwd}w{c}{\fonecolwd}w{c}{\emjudgecolwd}w{c}{\emjudgecolwd}}
\toprule
 & \multicolumn{6}{c}{\textbf{Answer $F_1$} $\uparrow$} & \multicolumn{1}{c}{EM $\uparrow$}
 & \multicolumn{1}{c}{Judge $\uparrow$} \\
\cmidrule(lr){2-7}\cmidrule(lr){8-8}\cmidrule(lr){9-9}
Stage & NQ & SQuAD & TriviaQA & EntityQ & WebQ & Mean & Mean & Mean \\
\midrule
SFT anchor, $0$ RL steps & 0.5412 & 0.6063 & 0.7176 & \textbf{0.5870} & 0.4779 & 0.5860 & 0.4810 & 0.6727 \\
\quad$+$ RL, step $20$ (early) & 0.5402 & 0.6113 & 0.7184 & 0.5858 & \textbf{0.4819} & 0.5875 & \textbf{0.4813} & 0.6779 \\
\quad$+$ RL, step $80$ (mid) & \textbf{0.5492} & \textbf{0.6318} & \textbf{0.7198} & 0.5817 & 0.4775 & \textbf{0.5920} & 0.4713 & \textbf{0.6925} \\
\quad$+$ RL, step $300$ (late) & 0.1156 & 0.1619 & 0.1439 & 0.1043 & 0.1121 & 0.1276 & 0.0000 & 0.5845 \\
\midrule
\lodestaricon~\ourstab{} (\fresp{}) & 0.4783 & 0.5916 & 0.7163 & 0.4943 & 0.3908 & 0.5343 & 0.4148 & 0.6420 \\
\bottomrule
\end{tabular}
\end{adjustbox}
\end{table}

\paragraph{Why the optimum is format-degenerate.}
The collapse is not an instability the run fell into; it is the objective being maximised.
Written out, the reward optimised here is
\begin{equation}
\begin{aligned}
R_{\mathrm{hybrid}}(y) &= R'_{\mathrm{CLR}}(y)\cdot\mathbf{1}(y^{*}\subseteq y),
&\qquad
R_{\mathrm{CLR}}(y) &= \mathcal{E}(y)\,\mathbf{1}\big(\mathcal{E}(y)>\tau\big)\big/\sqrt{T},\\
\mathcal{E}(y) &= S(y\mid D)-\min_{d\in D^{+}} S(y\mid D\setminus\{d\}),
&\qquad
S(y\mid\cdot) &= \textstyle\sum_{t=1}^{T}\log P(y_t\mid y_{<t},q,\cdot),
\end{aligned}
\label{eq:ctrlrag-reward}
\end{equation}
with $T$ the answer's token count and $\tau=1$. Two properties of that form push the same way
on $T$. First, $\mathcal{E}(y)$ is a difference of two sums over the \emph{same} $T$ tokens, so
it is extensive in length: each additional token contributes one more per-token contrast, while
the normaliser grows only as $\sqrt{T}$. Any positive average per-token contrast $c$ therefore
leaves $R_{\mathrm{CLR}}\approx c\sqrt{T}$, rising without bound in $T$\textemdash{}a sub-linear
normaliser does not cancel an extensive numerator, it only slows it, and the threshold pushes
the same way because a longer $y$ clears $\mathcal{E}(y)>\tau$ more easily. Second, the accuracy
factor is a \emph{containment} test, and containment is monotone under appending: once $y^{*}$
occurs in $y$, no continuation can remove it. Padding is thus free on the factor that checks
correctness and strictly profitable on the factor that does not, and the in-group min-max
rescale is order-preserving, so in every group the longer surviving sibling carries the larger
advantage. The joint optimum is to emit the gold string and then continue to the cap, which is
exactly the endpoint Table~\ref{tab:ctrlrag} records: exact match $0.0000$ while the judge still
credits $0.5845$, i.e.\ $y^{*}$ is still there and everything after it is reward being
collected. The analysis fixes the mechanism; Figure~\ref{fig:ctrlrag-reward} fixes its sign,
the reward rising over precisely the steps in which the responses lengthen. Note that this
argument is a property of the reward's shape alone, so the frozen-scorer deviation of
Appendix~\ref{app:baselines} does not bear on it.

\paragraph{Scoped conclusion.}
Our claim is bounded by our adaptation, which Appendix~\ref{app:baselines} discloses in
full: \emph{under that adaptation}, the contrastive-likelihood objective's optimum is
format-degenerate, the SFT anchor delays that optimum in proportion to its strength
without preventing it, and pre-declared cleaning rules are evaded by the next variant.
We explicitly do \emph{not} claim this of CTRL-RAG at its published scale, where the
official $74$k SFT corpus, a batch of $1024$ and a reward that reads likelihoods from
the policy itself rather than from a frozen copy are all untested here. The length
incentive we audited exists under any conditional likelihood, but its interaction with a
larger anchor and a smaller gradient noise scale is an open question.

\section{Baseline configurations and disclosures}
\label{app:baselines}

This appendix gives the per-row configuration behind the bare method names of
Table~\ref{tab:main}, together with the disclosures about how those
rows should be read.
Every configuration follows the fairness contract of the
\emph{Baselines} paragraph (Section~\ref{sec:settings}), and every method-by-dataset
cell of the \texttt{entropy}, \texttt{rank1} and \texttt{oracle} columns is identical
across all runs, confirming that every method ran on exactly the same items. The three
reference lines are \texttt{random}, the expected $F_1$ of a uniform draw;
\texttt{rank1}, the retriever's own top-ranked passage; and \texttt{oracle}, the best of
the ten in hindsight.

GainRAG, CRITIC-R1 and MBA-RAG are the only baselines that need a training artifact of
their own, so they are the rows for which the paper's own numbers cannot answer whether we
ran the other method correctly or ran it badly in our own favour. We therefore re-ran all
three from their stored artifacts on the WebQuestions pool, on different hardware from the
original runs, and all three return their tabulated value exactly ($0.3753$, $0.3674$ and
$0.3491$; $\Delta F_1 = 0.0000$ in each case). MBA-RAG's re-run coincides with the
\texttt{rank1} line because its bandit selects rank~1 on all $1{,}000$ questions of that
pool, which is that policy's collapse onto the retrieval prior reproducing rather
than a failed re-run.

\subsection{Training-free selection signals}
These rows score each candidate from the frozen respondent's own response and select by
that score, except for \texttt{SPS}, which scores the passage representation alone, and
Self-RAG (reflection), which scores with the method's officially released critic. None of
them is trained for this paper. Each row's configuration and the disclosure attaching to it are given below, one method
per entry, in Table~\ref{tab:main}'s order. Multi-sample methods use a uniform
$k\!=\!5$ budget unless an official row is given.
\begin{itemize}
\itemsep2pt
\item \textbf{Semantic entropy} \citep{farquhar2024semantic} runs in its official strict-entailment form. Its row is
inflated by the table's first-wins tie-break and should not be read as second place:
under that setting it ties on $62$--$79\%$ of questions with only $\sim\!2$ distinct
scores per $10$ candidates, so first-wins silently imports a retrieval-rank-1 prior.
Breaking ties at random gives mean $F_1$ $0.4774$, statistically indistinguishable from
\texttt{rank1}'s $0.4769$.
Official code at \url{https://github.com/jlko/semantic_uncertainty}.

\item \textbf{Self-RAG (reflection)} \citep{asai2024selfrag} selects with the method's officially released
critic. We score it over the full vocabulary instead of the official top-$5000$
log-prob truncation, a deviation strictly favourable to it.
Official code at \url{https://github.com/AkariAsai/self-rag}.

\item \textbf{SeaKR} \citep{yao2025seakr} runs at its official $n\!=\!20$. Our row omits the official newline
stop-strings, fixed seed and empty-sample filtering; restoring all three moves mean
$F_1$ by $+0.0006$, mixed in sign per dataset, and leaves every per-dataset comparison
against \ours{} unchanged.
Official code at \url{https://github.com/THU-KEG/SeaKR}.

\item \textbf{CLeHe} \citep{clehe2025} is our \emph{adaptation} of that method's entropy signal into a
selector. The official method is a token-level soft fusion that never selects a single
passage, so this row carries no inferential value about the published method.
Official code at \url{https://github.com/zexuanqiu/entropy-based-decoding}.

\item \textbf{EPR} \citep{epr2026} runs its official sequence-mean rate. Our row generates $24$ tokens
where the official script uses $200$; a rerun at the official budget moves mean $F_1$ by
$-0.0012$, so the row we rank against is the marginally \emph{stronger} configuration.
Official code at \url{https://github.com/artefactory/artefactual}, release tag
\texttt{ECIR2026}.

\item \textbf{Min-K\% Prob} \citep{shi2024mink} runs at its official $k\!=\!20$, adapted to selection by
scoring each passage's own cached greedy answer under the shared answering prompt.
Official code at \url{https://github.com/swj0419/detect-pretrain-code}.

\item \textbf{IGP} \citep{song2026igp} runs under its \emph{published} configuration, which probes with its
own short prompt rather than the answering prompt. That misalignment, not the gain
formulation, is the likely main cause of its position (Section~\ref{sec:related}); the
aligned-probe ablation of Appendix~\ref{app:alignedprobe} recovers about half the gap.

\item \textbf{EigenScore} \citep{chen2024inside} runs the released code at $k\!=\!5$ ($K\!=\!10$, $T\!=\!0.5$,
\texttt{max\_new\_tokens} $=\!256$) without Feature Clipping, which the official
repository also omits, so it corresponds to that paper's ``EigenScore (w/o)'' variant.
The released code also follows top-$k\!=\!10$ where the paper states $5$, and pools at
the EOS position.
Official code at \url{https://github.com/alibaba/eigenscore}.

\item \textbf{SPS} \citep{sps2026} runs its released scoring function, which reads the candidate text and never the
question. In SPS's own protocol the candidates are query-conditioned rewrites of one
evidence set, so nothing is lost by this; ours are ten \emph{different} passages, and a
score blind to the question cannot prefer the one that answers it. Its low row is that
mismatch, which we state before reporting it, and we keep the row beside the
\texttt{random} reference line so its position is checkable rather than asserted: its
correlation with per-candidate answer $F_1$ is $-0.005$ to $-0.082$ on our pools, its mean
$F_1$ $0.3380$ sits at that line's $0.3271$ (within $-0.5$ to $+2.4$ paired $z$ per
dataset), the implementation was verified against the released code element by element
(subspace basis, PCA rule, max-pooled penultimate representation, bare-sentence input,
residual score, argmin), and the argmax control that would have exposed a sign error is
worse on all five datasets.
Official code at \url{https://github.com/HU-xiaobai/Code-of-Beyond-Perplexity-Let-the-Reader-Select-Retrieval-Summaries-via-Spectrum-Projection-Score}.

\item \textbf{Min-K\%++} \citep{zhang2025minkpp} runs the identical transplant, differing
only in its vocabulary-normalized per-token statistic. Its paper fixes no default $k$, so
the run also sweeps $k{=}10..100$, and the best point ($k{=}10$, macro $0.2912$) still
sits below every other ranked row. Its below-\texttt{random} row is a property of the
statistic under this transplant, not an implementation defect: our reimplementation
matches the official release term for term, and an argmin control does not rescue the
score.
Official code at \url{https://github.com/zjysteven/mink-plus-plus}.
\end{itemize}
Every ranked row with tie logging other than semantic entropy has a tie rate
$\leq 1.2\%$ and reproduces its first-wins numbers under random tie-breaking. Plain
first-token entropy, the polarizer ablation of \ours{}, is not ranked as a baseline row
of Tables~\ref{tab:main} and~\ref{tab:em}; being our own method minus one component, it
is reported with the ablations (Section~\ref{sec:ablation}) and paired-tested in
Table~\ref{tab:sig}. At mean $F_1$ $0.5148$ it is nonetheless the strongest \emph{untrained}
selector in the benchmark, behind only the searched GEPA string among inference-ready configurations, so
what \ours{} improves on is a demanding comparison rather than a favourable one.

\subsection{Searched guidance}
The prompt-optimization rows ask whether \ours{}'s GRPO stage can be replaced by searching the
polarizer directly. Both configurations consume \ours{}'s exact reward function, training draws and frozen
respondent, search under the same $96$-token polarizer cap, and select identically at inference
($\arg\min_p H_1(q,p,\psi)$); only the optimizer differs. Both also start from the \emph{same
hand-written caution seed}, a manually written polarizer directing doubt at passages that look
relevant but do not answer the exact question, and on the never-searched pools the searched
strings improve on it by at most $+0.003$ mean $F_1$. That seed is itself evaluated under the
identical protocol and reaches mean $F_1$ $0.5129$ and EM $0.4005$, \emph{below} the
no-polarizer ablation's $0.5148$: writing the polarizer by hand adds nothing over inserting no
polarizer at all, so the ordering across the three ways of obtaining the string is hand-written
$0.5129$, searched $0.5131$--$0.5160$, GRPO-learned $0.5339$. Each optimizer's configuration is
given below, one per entry, in Table~\ref{tab:main}'s order.
\begin{itemize}
\itemsep2pt
\item \textbf{GEPA} \citep{agrawal2025gepa} runs its official implementation, reflective
evolution with Pareto candidate selection and gpt-oss-120b as the reflection LM, at two
settings. At its official sample-efficiency setting, all six runs we launched\textemdash{}three
seeds at each of two reflection-minibatch sizes\textemdash{}returned the hand-written seed
unchanged, and not for want of proposals: at GEPA's default minibatch of three, $23$--$25$ of
the reflection LM's rewrites per run cleared the minibatch screen and were fully validated
without ever displacing the seed on the Pareto frontier, and at a minibatch of ten no rewrite
cleared the screen at all. The second setting runs each seed to saturation under plateau
early-stopping; the table reports the strongest saturated seed.

\item \textbf{MIPROv2} \citep{opsahlong2024mipro}, a non-reflective Bayesian instruction search,
runs under the same harness at the same sample-efficiency setting. Unlike GEPA there, every
seed improved on the hand-written seed internally ($+0.005$ to $+0.107$ separation)\textemdash{}the
same harness accepting improvements whenever the optimizer finds one, which rules out a harness
fault behind GEPA's unchanged returns\textemdash{}and the table reports the strongest seed.
\end{itemize}
Each row therefore prints its strongest seed against \ours{}'s three-seed mean, a
comparison deliberately set in the baselines' favour. On the never-searched pools, however,
both searched strings land within $+0.003$ mean
$F_1$ of the untuned hand-written caution\textemdash{}and significantly below \ours{}
(Table~\ref{tab:sig}). Section~\ref{sec:main} reports the full comparison.

\subsection{Trained selectors}
These three rows replace the selection signal with a model trained on our data, while the
same frozen respondent still answers.

\begin{itemize}
\itemsep2pt
\item \textbf{GainRAG} \citep{gainrag2025} is the only baseline that, like \ours{}, \emph{trains}
on our data, so we control its training distribution exactly. It enters only as
its NQ-trained \emph{inference-ready} distilled selector: its per-candidate gain signal
reads the gold answer and is therefore not a selector one can run at inference, so we
report that signal at the end of this entry rather than as a table row.
Following its official
recipe, per-candidate gains are synthesised as contrastive-decoding perplexities of
the gold answer ($\alpha{=}0.5$) under the same frozen respondent and the same answering
prompt that produced the cached utilities; a respondent-generated pseudo-passage joins
the ten bge-m3 candidates ($11$ per question); questions whose highest-gain candidate
fails GainRAG's lenient containment-EM filter are dropped; and bge-reranker-base is
distillation-finetuned on the soft gain distribution (KL loss on
$-\log(\mathrm{PPL}{+}1)$ targets, group size $11$, lr $6\times10^{-5}$, $2$ epochs,
official defaults otherwise, final checkpoint, no model selection). Crucially, its
training questions are \emph{exactly} the $1{,}939$ NQ-open train questions \ours{}
trains on (the filter keeps $1{,}883$), so both methods see the same supervision
distribution and share an in-domain pool. Two benchmark adaptations from the paper's
setting: candidates come from the shared bge-m3 top-$10$ pools rather than Contriever
top-$20$, and reranker input length is capped at the model's true $512$-token limit
rather than the released script's $512{+}512$. At test time the selector scores the same
ten candidates as every other row, one $278$M cross-encoder forward per candidate with
no respondent and no gold; its full workflow with the pseudo-passage action enabled is
not cell-comparable and is excluded from the tables, where it would reach mean $F_1$
$0.4907$.
The \emph{teacher} signal is likewise kept out of the tables, but we report it because
it bounds what the distillation could have delivered\textemdash{}selecting by a gain
that scores the gold answer is selecting with the answer in hand. Reproduced on our
pools it reaches mean $F_1$ $0.6213$ over the four datasets where it is available,
against \ours{}'s $0.4888$ on those same four, while the distilled selector retains
$0.4480$ on those four ($0.4951$ over all five pools, its Table~\ref{tab:main} row).
That gap is distillation loss rather than a reproduction failure, and it is why
GainRAG's inference-ready row sits mid-table.

\item \textbf{MBA-RAG} \citep{tang2025mbarag} trains a DistilBERT bandit policy with an
$\epsilon$-greedy reward $r_a = A(y, \hat{y}_a) - \lambda C(a)$. Its published arms
are retrieval \emph{strategies}; our benchmark fixes retrieval, so the arm set is
adapted to the ten candidate ranks of the shared pools, and since every arm then costs
exactly one respondent call the cost term is degenerate and $\lambda$ is set to $0$.
Rewards are the same cached per-passage $F_1$ utilities that define our labels.
\emph{Result:} the trained policy collapses onto the retrieval prior, reaching mean
$F_1$ $0.4765$; the paired comparison against \ours{} is in Table~\ref{tab:sig}.

\item \textbf{CRITIC-R1} \citep{criticr12026} trains a structured critic
(verdict/location/reason/fix schema) with two-stage GRPO under consensus
teacher supervision. We adapt it to selection: the trained critic scores each
candidate's cached draft answer by its verdict-token probabilities
($P(\texttt{CORRECT})-P(\texttt{INCORRECT})$, one forward per candidate), the
best-scored passage is selected, and the same frozen respondent answers. Training
follows the official recipe at the official scale: Qwen2.5-3B-Instruct
critic with LoRA \citep{hu2022lora} $r{=}16/\alpha{=}16$, lr $10^{-6}$, KL $0.003$, $5{,}000$
critique samples at the official five-epoch exposure, two stages (CJA then gated
DQA) with the paper's conservative verdict matrix, the training
triples drawn only from \ours{}'s train questions and consensus supervision
($K{=}3$) from a gpt-oss-120b teacher over the same cached draft answers.
\emph{Result:} the critic trains healthily by its own reward yet its selection
ends at mean $F_1$ $0.4626$, below \texttt{rank1}'s $0.4769$; the paired comparison
against \ours{} is in Table~\ref{tab:sig}. An implementation audit accompanies the
number, and the residual gap is the method rather than the harness: teacher-verdict
alignment and within-question ranking are decoupled skills on this benchmark, and the
two training stages reward the former.
\end{itemize}

\subsection{Trained generators (system-level)}
In these two rows the answering model itself is trained, so they are
\emph{system-level} comparisons rather than cell-level ones and are grouped
separately in the tables.
\begin{itemize}
\itemsep2pt
\item \textbf{Self-RAG} \citep{asai2024selfrag} is the released
\texttt{selfrag-llama2-7b} generator running its official reflection-token
inference over the same ten candidates. Unlike every other row it replaces the
frozen respondent with a trained one, a Llama-2-7B fine-tuned on Self-RAG's own
$150$k-instance corpus, so its row answers the question ``does a purpose-trained
retrieval-augmented generator beat a frozen respondent steered by one learned
string?''
\emph{Result and metric diagnostic.} Its reflection-based passage selection is
sound (frozen-respondent $F_1$ at its chosen passages matches plain entropy within
$\pm 0.015$ on every pool), but its answers are style-mismatched with token-level
$F_1$: on well-formed out-of-domain questions it answers in full sentences (mean
$11.9$ words vs.\ the frozen respondent's $4.2$), which token $F_1$ punishes even when
the fact is present. Its table row therefore uses its stronger configuration
under our one-row policy, official inference plus a gold-agnostic
extraction step, which recovers mean $F_1$ from $0.3346$ to $0.3638$
(Table~\ref{tab:trainedgen}); \ours{} loses only the NQ cell to it, by $0.001$, consistent
with NQ-style data in its training corpus.
Under Self-RAG's \emph{own} official metric, containment match, the comparison is a
statistical tie, because containment rewards its verbose style; we report both metrics
rather than choosing either.
\item \textbf{CTRL-RAG} \citep{ctrlrag2026} also trains the generator itself rather than a
selector: full-parameter GRPO on the very Llama-3.1-8B-Instruct checkpoint we
otherwise keep frozen, optimizing the paper's contrastive-likelihood hybrid reward
$R_{\mathrm{hybrid}} = R'_{\mathrm{CLR}} \cdot \mathbf{1}(y^{*} \subseteq y)$,
whose leave-one-out likelihood gap is scored online by a frozen-likelihood service
held at the initial policy. We keep its two-stage SFT-then-RL structure but rebuild
the SFT stage inside the benchmark, on $1{,}839$ of the $1{,}939$ NQ training questions
\ours{}
trains on, with the first gold answer as the target, in place of the official $74$k
multi-hop SFT corpus, which would inject supervision from outside this comparison. It
reads the shared top-$10$ concatenation and answers directly, with no selection step,
so its cells are not frozen-respondent-comparable and are not claimed in the per-cell
win count. It is kept out of Tables~\ref{tab:main}, \ref{tab:sig} and~\ref{tab:em}, and
appears only beside Self-RAG in Table~\ref{tab:trainedgen}; its scores, formatting
pathology and the reason no single checkpoint of the run can stand in for the method are
reported together in the reproduction case study of Appendix~\ref{sec:ctrlrag}.
\end{itemize}

\end{document}